\documentclass[preprint]{elsarticle}

\AtBeginDocument{%
  \providecommand\BibTeX{{%
    \normalfont B\kern-0.5em{\scshape i\kern-0.25em b}\kern-0.8em\TeX}}}

\usepackage{algorithmic}
\usepackage{textcomp}
\usepackage{graphicx}
\usepackage{xcolor}
\usepackage{multirow,makecell}
\usepackage{array}
\usepackage{longtable}
\usepackage{pdflscape}
\usepackage{amsmath}
\DeclareMathOperator*{\argmax}{arg\,max}

\begin{document}

\title{Towards a Robust and Trustworthy Machine Learning System Development: An Engineering Perspective}

\author[inst]{Pulei Xiong\footnote{Corresponding author}, Scott Buffett\footnote{These authors contributed equally to this research}, Shahrear Iqbal\textsuperscript{2}, Philippe Lamontagne\textsuperscript{2}, Mohammad Mamun\textsuperscript{2}, Heather Molyneaux\textsuperscript{2}}

\affiliation[inst]{organization={Cybersecurity, National Research Council Canada},
            addressline={1200 Montreal Rd}, 
            city={Ottawa},
            postcode={K1A 0R6}, 
            state={ON},
            country={CA}}

\begin{abstract}
While Machine Learning (ML) technologies are widely adopted in many mission critical fields to support intelligent decision-making, concerns remain about  system resilience against ML-specific security attacks and privacy breaches as well as the trust that users have in these systems. In this article, we present our recent systematic and comprehensive survey on the state-of-the-art ML robustness and trustworthiness from a \emph{security engineering} perspective, focusing on the problems in system threat analysis, design and evaluation faced in developing practical machine learning applications, in terms of robustness and user trust. Accordingly, we organize the presentation of this survey intended to facilitate the convey of the body of knowledge from this angle. We then describe a metamodel we created that represents the body of knowledge in a standard and visualized way. We further illustrate how to leverage the metamodel to guide a systematic threat analysis and security design process which extends and scales up the classic process. Finally, we propose the future research directions motivated by our findings. Our work differs itself from the existing surveys by (i) exploring the fundamental principles and best practices to support robust and trustworthy ML system development, and (ii) studying the interplay of robustness and user trust in the context of ML systems. We expect this survey provides a big picture for machine learning security practitioners. 
\end{abstract}


\maketitle

\section{Introduction}

In recent years, Machine Learning technologies, especially the Artificial Deep Neural Networks (DNNs) and Deep Learning (DL) architectures, have been widely adopted in many mission critical fields, such as cyber security, autonomous vehicle control, healthcare, etc. to support intelligent decision-making~\cite{8779838}. While ML has demonstrated impressive performance over conventional methods in these applications, concerns exist regarding system resilience against ML-specific security attacks and privacy breaches as well as the trust that users have in these systems~\cite{LecunBengioBengio,BIGGIO2018317,PrithvirajJoseph}.

With the impressive success of applying ML in various application areas, security weaknesses inherent in ML technologies, e.g., learning algorithms or generated models, have been revealed by a large number of researchers~\cite{8779838,BIGGIO2018317}. Due to these weaknesses,  ML systems are vulnerable to various types of adversarial exploitations that can compromise the entire system. In fact, a typical ML pipeline, which consists of data collection, feature extraction, model training, prediction, and model re-training, is vulnerable to malicious attacks at every phase~\cite{WANG201912}. The attacks against ML systems have negative impacts on the systems that may result in performance decrease, system misbehavior, and/or privacy breach~\cite{PrithvirajJoseph, 8677282}. Machine learning and cyber security researchers are greatly motivated to uncover these ML inherent weaknesses, exploitable vulnerabilities and applicable attacks, and have been working hard to develop effective defense mechanisms.

The development of robust and trustworthy ML systems is a multi-disciplinary endeavour spanning machine learning, cyber security, human-computer interaction, and domain-specific knowledge. The \emph{robustness} of an ML system can be defined as its resilience to malicious attacks to protect itself from the compromise of the system’s integrity, availability, and confidentiality. A robust ML system can inspire \emph{user trust} in the system’s security compliance, while users' trust in an ML system can assist in achieving a system's security objectives by helping users to take appropriate responses to malicious attacks and to avoid incidental actions.

The ML/AI community recognizes that all-hands efforts at various levels are needed to support and ensure the development of robust and trustworthy ML systems. Policymakers around the world have made a number of ongoing efforts on regulation enactment to support and normalize AI practitioners' behaviors~\cite{TheLawLibraryofCongress2019}. For instance, the Government of Canada is developing the Algorithmic Impact Assessment (AIA)\footnote{https://www.canada.ca/en/government/system/digital-government/digital-government-innovations/responsible-use-ai/algorithmic-impact-assessment.html; visited on 11/24/2020} under the Directive on Automated Decision-Making\footnote{https://www.tbs-sct.gc.ca/pol/doc-eng.aspx?id=32592; visited on 11/24/2020}. AIA is an online questionnaire tool designed to help identify the impact level of an automated decision system. Over 80 organizations in both public and private sectors have taken the step to develop AI ethic principles to guide responsible AI development, deployment, and governance~\cite{Mittelstadt2019}. A recent report, ``Toward Trustworthy AI Development: Mechanisms for Supporting Verifiable Claims"~\cite{Brundage2020TowardTA}, represents a joint effort of academia and industry to move beyond the ethic principles by proposing a set of mechanisms that AI practitioners can adopt to make and verify claims about AI systems. These verifiable claims, as evidence for demonstrating responsible behavior, can be used to enforce the compliance of the regulations and norms mandated in the high-level AI ethical principles. 

\subsection{Our Contributions and Related Work}\mbox{}\\
This work presents our contributions from a security engineering perspective to the development of robust and trustworthy ML systems. From the security engineering perspective, the system risks introduced by adopting ML technologies should be considered at the design level when an ML-powered system is developed. This discipline is also known as the proactive, security-by-design approach \cite{8779838,BIGGIO2018317}. In the context of ML technologies, it consists of the process including : (i) understanding the inherent weakness in ML techniques; (ii) developing a threat model for an ML system; (iii) identifying potential ML vulnerabilities and attack vectors; (iv) designing appropriate countermeasures; and (v) applying a systematic evaluation methodology for the ML system security assurance and compliance.

We conducted a systematic and comprehensive survey on the state-of-the-art robustness and trustworthiness technologies for machine learning systems. The scope of this survey is to cover the literature published in peer reviewed conferences and journals with the focus on the work published since year 2014 (five years before we started our survey). During the review, we consciously traced the trend in the area and included the newly published papers that are most relevant. We also carefully expanded the scope to include the literature that was published before 2014 but has significant influence and the papers published on arXiv~\footnote{https://arxiv.org/; visited on 11/10/2021} only (no peer-reviewed yet) but provide the most recent progress in the area.

\begin{figure}[htbp]
\centerline{\includegraphics[width=0.75\textwidth]{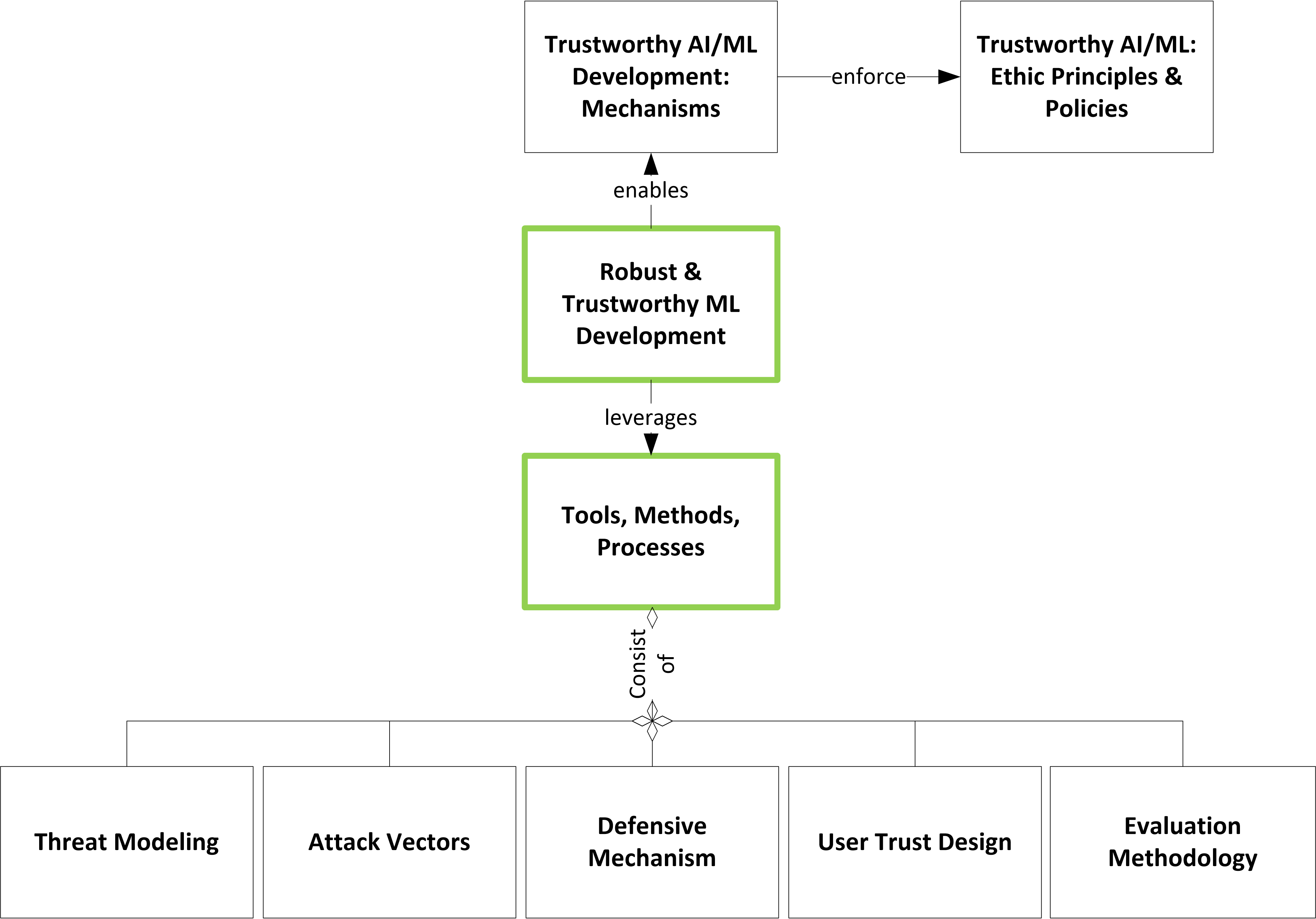}}
\caption{Robust and Trustworthy ML System: An Ecosystem View}
\label{fig:Ecosystem}
\end{figure}

We then pushed our effort forward above and beyond a survey by developing a metamodel specified in Unified Modeling Language (UML)\footnote{uml.org; visited on 09/28/2020}, which captures and represents the body of knowledge in a standard and visualized way. We further studied how the metamodel can be used to guide a systematic process to perform threat analysis and security design in the ML system development, which extends and scales up the classic process. Figure \ref{fig:Ecosystem} depicts the all-hands efforts at various levels needed for robust and trustworthy ML system development, and the focuses of our effort (the bold, green boxes). To the best of our knowledge, our work is the first of its kind of engineering effort to address the gap of knowledge in ML system development.

Our work differentiates itself from the existing survey papers in the area in two important aspects: (i) we explore the fundamental principles and best practices to support robust and trustworthy ML system development; and (ii) we study the interplay of robustness and user trust in the context of ML systems. The existing surveys, including~\cite{BIGGIO2018317,WANG201912,barreno2010security,Xue2020,Papernot2018,Liu2018,Ren2020,Zhang2020,He2019,Miller2020,Serban2020,Yuan2019,Qiu2019,Akhtar2018,Ozdag2018,Holt2017,Pitropakis2019,Gardiner:2016:SML:2988524.3003816,Dasgupta2019}, primarily focused only on ML defensive and offensive technologies: in~\cite{Xue2020,WANG201912,BIGGIO2018317,Papernot2018,Liu2018,barreno2010security}, the authors presented a comprehensive robust ML offensive and defensive technologies including threat modeling and evaluation methods; in~\cite{Ren2020,Zhang2020,He2019,Miller2020,Serban2020,Yuan2019,Qiu2019,Akhtar2018,Ozdag2018,Holt2017}, the authors focused their investigation on the attack and defense methods against DL/DNNs, with (i) a survey dedicated on deep learning in computer vision~\cite{Akhtar2018}, (ii) a survey focused particularly on defenses mechanisms~\cite{Miller2020}, and (iii) the analysis of the NIPS 2017 Adversarial Learning Competition results~\cite{Ozdag2018}; in~\cite{Pitropakis2019}, Pitropakis {\em et al.} dedicated their efforts on the taxonomy and analysis of attack vectors against plenty varieties of ML algorithms in a broad range of ML applications; in~\cite{Gardiner:2016:SML:2988524.3003816}, Gardiner and Nagaraja investigated attacks against various supervised and unsupervised learning algorithms used in malware C\&C detection; in~\cite{Dasgupta2019}, Dasgupta {\em et al.} conducted a detailed survey on the robust ML techniques by using the computational framework of game theory.

In consideration of the scope, the engineering angle, and the suitable length for this survey, we made a trade-off between the level of detail and the diversity of topics to present. We choose the topics that are directly relevant to the context of robust and user trust machine learning system development, and present them at the level of detail that describes the overall landscape with important concepts, available methods, applicable process, etc. The readers are encouraged to read the original literature when they need to dig into the content such as the details of algorithms, experiment results, application scenarios. 

\subsection{Organization of the Article}\mbox{}\\
The rest of this paper is organized as follows. Sec~\ref{section:ML Overview} sets up the context of ML technologies discussed in this article. Sec~\ref{section:Threat Modeling} to~\ref{section:Assessment and Metrics} summarize our findings in the literature on the key technologies in supporting robust and trustworthy ML system development, including threat modeling, common offensive and defensive technologies, privacy-preserving machine learning, user trust in the context of machine learning, and empirical evaluation for ML model robustness. Sec~\ref{section:ML Development} describes a metamodel we created that represents the body of knowledge we learned from the survey, and then illustrates a systematic approach guided by the metamodel to performing threat analysis and security design for ML systems. Sec~\ref{section:Conclusions} concludes our work and proposes future research directions motivated by our findings to advance the development of robust and trustworthy ML systems.

\section{Secure Machine Learning: An Overview}
\label{section:ML Overview}

Machine learning encompasses a variety of approaches that facilitate problem solving through experience~\cite{mitchell1997machine, mitchell1999machine}, typically by enabling the discovery of important patterns or regularities in large datasets. Machine Learning approaches can broadly be classified into three major paradigms: supervised learning, unsupervised learning and reinforcement learning. Each of these paradigms exhibits their own vulnerabilities. In this section, an overview is provided for each paradigm with introductions to the relevant techniques and models they include, followed by an introduction of some of the potential vulnerabilities as well as a brief review of possible exploitations as documented in the literature.

With supervised learning techniques, the objective is to develop a function that can map input instances to labels, using a set of examples upon which to train a model. The idea here is that, given the assumption that the sample used for training is representative of the population, a function that can be derived to perform well at correctly labeling the training data should perform well at labeling new data. So-called discriminative modeling approaches, such as logistic regression and support-vector machines, can then be used to predict the likelihood of a new instance belonging to a particular class. This is done by determining a direct mapping from feature values to labels, for example by determining a boundary in the data separating the two (or more) classes. Conversely, generative modeling approaches, such as Naïve Bayes Classification, use the probabilities of the feature values that make up an example instance to compute the likelihood of each class. Artificial neural network-based approaches, such as deep learning, can also be used in a supervised manner to learn high-level features, such as those required for image processing, but can also be utilized in a semi-supervised or unsupervised manner. Few-shot learning approaches~\cite{wang2020generalizing} can be leveraged when there are relatively few examples upon which to learn a classification model, while zero-shot learning approaches~\cite{wang2019survey} are applicable when instances to be classified might belong to classes that are not seen during training.

Rather than relying on a set of examples upon which to train a classifier, unsupervised machine learning approaches instead look for other similarities in the data that can be exploited in such a way as to make possible inferences or assumptions during learning and prediction. Clustering methods focus on identifying certain commonalities among the data, which can then be used to make assertions about certain data depending on the level of fit. Anomaly detection, for example, can be used to deem particular instances as abnormal, providing evidence that they may be of particular interest. Malicious network behaviour, as an example, might be identified using unsupervised anomaly detection approaches that can identify patterns that are inconsistent with typical observed activity.

Reinforcement learning is alternate paradigm where learning is conducted in an exploratory manner, often modeled by a Markov Decision Process. The objective is to learn a solution to a problem, where proposed solutions can be evaluated via a reward function. Learning thus modifies solutions while seeking to maximize rewards.

The ubiquitous nature of machine learning techniques, and their subsequent rapid adoption, has resulted in increased vulnerability of systems and greater attractiveness for potential attackers~\cite{pitropakis2019taxonomy}. Potential network attackers may want to influence a ML-based Intrusion Detection System (IDS) to increase false negatives, allowing the attackers to enter undetected, or increase false positives to the point that so much legitimate traffic is flagged that alerts become too frequent. In such a case, either they are ignored, or operation is disrupted altogether via denial of service (DoS)~\cite{barreno2010security}. Advertisers may similarly seek to influence spam detectors to increase the likelihood of their messages penetrating email filters~\cite{lowd2005adversarial}. Training data for image recognition may be perturbed in such a way to allow unauthorized access or cause harm in other domains, such as connected and automated vehicles~\cite{taeihagh2019governing}.

To illustrate a general model of security for supervised machine learning, Barreno {\em et al.}~\cite{barreno2010security} offered a taxonomy that divides the aspects of vulnerabilities and attacks along three different dimensions. Their discussion is framed in the context of a machine learning system that is designed to identify and defend towards potential attackers, but is generalizable to supervised ML systems. The three dimensions are as follows:

\begin{itemize}
\item {\em Influence: Causative vs Explorative}. This indicates whether the training data is compromised (causative), resulting in a faulty prediction or classification model being produced, or the classification of new data itself is compromised (explorative) in real time. Pitropakis {\em et al.}~\cite{pitropakis2019taxonomy} refer to this as {\em poisoning} vs {\em evasion}.
\item {\em Security Violation: Integrity vs Availability}. This dictates whether the exploitation focuses on compromise via the generation of false negatives (integrity), or via overload of false positives (availability).
\item {\em Specificity: Targeted vs Indiscriminate}. This aspect pertains to whether a particular instance is the focus (targeted) or a wider class (indiscriminate). 
\end{itemize}

There is a significant body of work that has explored various vulnerabilities of machine learning systems and how they might be exploited. Barreno {\em et al.}~\cite{barreno2010security} center their discussion within the context of attacks on intrusion detectors, but also offer a detailed illustration on how attacks on a spam filter could fit within each possible outcome for the aforementioned taxonomy. Yuan {\em et al.}~\cite{yuan2019adversarial} explored attack mechanisms on deep learning systems, specifically at the classification/validation stage, as opposed to the training stage, positioning such attacks as explorative/evasion in the Barreno taxonomy. Su {\em et al.}~\cite{su2019one} described a similar deep neural network vulnerability, and illustrate how image classification can be drastically modified via the perturbation of just a single pixel, facilitating attacks that would fall into the indiscriminate class of the taxonomy. Pitropakis {\em et al.}~\cite{pitropakis2019taxonomy} further provided an extensive survey of ML vulnerabilities and associated attack strategies. An in-depth review of the more significant advances in this area is detailed throughout sections~\ref{section:Threat Modeling} to~\ref{section:ML Privacy}.

\section{Threat Modeling}
\label{section:Threat Modeling}
\emph{Threat Modeling} is an engineering technique to support systematic security requirement analysis. It has been widely adopted by cyber security researchers and professionals to identify potential system threats, set feasible security objectives, identify relevant vulnerabilities and attack vectors, and design appropriate defense mechanisms. A well-defined threat model serves as a backbone of the secure development process to reduce the risk of security issues arising during the application development and shape the application security design to meet the security objectives. 

In the context of ML security, the researchers focused on the following aspects of threat modeling~\cite{WANG201912,BIGGIO2018317,Papernot2018,Gardiner:2016:SML:2988524.3003816,Papernot2016TowardsTS}:

\paragraph{\textbf{Attack Surface}}\mbox{}\\ 
In Machine Learning, the workflow of the complete ML tasks is modelled as a \emph{Pipeline} which consists of several phases, including data collection, data pre-processing, feature extraction, model training and testing, prediction, and optionally model re-training. Tremendous sensitive and confidential data, from raw data to trained models, flows along the pipeline. A number of attack surface and various attack vectors have been identified in the pipeline as summarized below:
\begin{enumerate}
\item[$\bullet$] Stealthy Channel attack during raw data collection phase;
\item[$\bullet$] Mimicry and Poisoning attack against training and testing datasets;
\item[$\bullet$] Polymorphic/Metamorphic attack against feature extraction;
\item[$\bullet$] Gradient Descent attack against learning algorithms;
\item[$\bullet$] Evasion attack against trained models during prediction phase; 
\item[$\bullet$] Model Stealing against trained models; and
\item[$\bullet$] Poisoning Attack during model re-training phase.
\end{enumerate}
The majority of the research we have reviewed is focused on \emph{Poisoning Attack}, \emph{Gradient Descent Attack}, \emph{Evasion Attack}, and \emph{Model Stealing}. 

\paragraph{\textbf{Attacker's Goal}}\mbox{}\\
Attacker's adversarial goals can be categorized from the following three perspectives:
\begin{enumerate}
\item[$\bullet$] \emph{Security Violation}. With the classical CIA model (confidentiality, integrity, availability), an attacker may aim to undermine an ML system's functionality (integrity and availability), or to deduce sensitive information about the ML system (confidentiality, or privacy): (i) \emph{integrity violation} via false negative, e.g. evade a spam email detection system without compromising the normal system operation; (ii) \emph{availability violation} via overload of false positives, e.g. significantly degrade the accuracy of a spam email detection system so that its functionalities are not available to legitimate users; or (iii) \emph{privacy violation} by stealing sensitive or confidential information from an ML system, e.g. obtaining a trained ML model parameters or training data, by an unauthorized approach.
\item[$\bullet$] \emph{Attack Specificity}. An attacker may launch \emph{targeted attacks} against a specific ML algorithm or architecture, or launch  \emph{indiscriminate attacks} against any ML system.
\item[$\bullet$] \emph{Error Specificity}. In the context of ML classifier systems, an attacker may aim to fool the system to misclassify an input sample to a specific class (\emph{error-specific attacks}) or to any of the classes different from the right class (\emph{error-generic attacks}).
\end{enumerate}
An adversarial attack may present a goal of the combination of these different characters. For example, an adversary is motivated to launch attacks to evade a given spam detection system by crafting malicious emails based on the algorithms specifically optimized against the system. 

\paragraph{\textbf{Attacker's Knowledge}}\mbox{}\\
The data and information related to an ML system, including training data, feature set, learning algorithms and architecture, hyperparameters, objective function, and trained model parameters (weights), are considered sensitive or confidential. Depending on the level of access to these data and information, an attacker can launch different types of attacks, that is, black-box based attack, gray-box based attack, and white-box based attack. 
\begin{enumerate}
\item[$\bullet$] \emph{Perfect-knowledge (PK)}, white-box attacks: An attacker knows everything about a targeted ML system including training dataset, ML architecture, learning algorithms, trained model parameters, etc. This setting represents the worst-case attacking scenario.
\item[$\bullet$] \emph{Limited-knowledge (LK)}, gray-box attacks: An attacker has a portion of the knowledge about a targeted ML system. Typically, the attacker is assumed to know the feature set, the model architecture and the learning algorithms, but not the training data and the trained model parameters. The attacker may be able to collect a surrogate dataset from a similar source and get feedback/output from the ML system to acquire labels for the data, and then use the information to launch attacks further.
\item[$\bullet$] \emph{Zero-knowledge (ZK)}, black-box attacks: An attacker is assumed to not know any ``exact'' information about a targeted ML system. However, this setting implies that inevitably the attacker is able to acquire partial but inaccurate information about the targeted system. For example, the attacker may not know the exact training data or feature representation used for training an object-detection model in autonomous vehicle control, the kind of data - images of traffic road signs, and the features - image pixels, are known to everyone including the attacker. While the black-box setting does increase the threshold of exploitability, an ML system is still vulnerable to various attacks.
\end{enumerate}

\paragraph{\textbf{Attacker's Capability}}\mbox{}\\
It refers to which extent an attacker can access and manipulate training data or input samples, or observe the corresponding output of a trained model. The level of attacker's access and manipulation of the data includes \emph{read}, \emph{inject}, \emph{modify}, or logically \emph{corrupt} training data or input samples, in the order of the capacity from weak to strong. 

\paragraph{\textbf{Attacking Influence}}\mbox{}\\
The attacker's influence can be categorized as \emph{Causative} if the attacker can manipulate both training data and input samples during the ML training (offline or online) and prediction phases, or \emph{Exploratory} if the attacker can only manipulate input samples. Causative attack attempts to influence or corrupt the model under training. The goal of causative attack can be integrity violation that causes the model produce an adversary desired outputs (error-specific attack) as the adversary supplies the model with the crafted input samples, or availability violation due to the logically corrupted model. Exploratory attack does not tamper with the targeted model. The goal of exploratory attack can be integrity violation that causes the model produce incorrect outputs, or privacy violation that deduces sensitive or confidential information about the model and training dataset.

\paragraph{\textbf{Attacking Strategy}}\mbox{}\\
It refers to the systematic approach that an attacker is to take to optimize the attacking effort. For example, depending on how much knowledge $(\theta)$ about an ML system and how much capability of accessing and manipulating training and/or input data that an attacker may have to generate attack samples $({\cal D}'_c$), the attacker can use an \emph{objective function}, defined as ${\cal A}({\cal D}'_c, \theta)$, to measure attacking effectiveness and optimize attacking methods and algorithms~\cite{BIGGIO2018317}. Equation~(\ref{eq:attack strategy}) defines a high-level formulation of attacking strategies for various attack vectors such as evasion and poisoning attacks.
\begin{equation} \label{eq:attack strategy}
  {\cal D}_c^* \in \argmax_{{\cal D}'_c\in \Phi({\cal D}_c)} {\cal A}({\cal D}'_c, \theta)
\end{equation}

\paragraph{\textbf{Attacker's Role}}\mbox{}\\ 
In the context of \emph{Privacy Preserving Machine Learning} (PPML), there are three different roles involved in the ML pipeline~\cite{8677282}: (i) \emph{Input Party} who is the owner or contributor of training data; (ii) \emph{Computation Party} who performs model training; and (iii) \emph{Results Party} who submits input samples to a trained model and receives results. It is common that  the computation party and the results party are the same entity, while the input party is a different entity. For example, the input party can be individuals around the world, while the computation party and the results party is a company which collects data, trains an ML model, and runs the trained model for its business.

\section{Common Machine Learning Offensive and Defensive Technologies}
\label{section:ML Offensive and Defensive Technologies}
In this and the next section, we  present commonly used ML offensive and defensive technologies. The prevailing publications during the past five years primarily addressed the adversarial sampling problems at the ML training and prediction phases against DNN/DL based supervised learning algorithms in the application areas of image classification and anomaly detection. The content in this and the next section reflects this research trend. This section summarizes our findings related to the security violation/defense of system availability and integrity.

\subsection{Attack Vectors}
\label{section:Attack Vectors}
More than a decade ago, researchers started to gain awareness of ML security problems and relevant adversarial attacks against traditional, non-deep learning algorithms, such as linear classifier used in spam filtering and Support Vector Machine (SVM) based binary classifier used in malicious PDF detection~\cite{BIGGIO2018317,Liu2018}. With the advances in the study of deep neural networks and deep learning architecture as well as the increasing application in various areas such as computer vision and cyber security, researchers continue to uncover vulnerabilities existing in the DNN/DL algorithms and architectures. While this research  mainly focuses on supervised ML methods, unsupervised ML methods (e.g. clustering) are also vulnerable to adversarial attacks~\cite{BIGGIO2018317,WANG201912,Liu2018}. The attack vectors found in the literature we reviewed are summarized in Table~\ref{tabsec4:attackvectors}.

Adversarial attacks against ML systems exist at every phase of the ML pipeline, but attacks against the model training and prediction phase, including poisoning attacks, evasion attacks, and privacy attacks, received the most interest. The way to launch these attacks can be categorized as input manipulation, input extraction, training data manipulation, training data extraction, model manipulation, and model extraction~\cite{8779838}. In the rest of this and the next section, these attack vectors will be discussed with respect to \emph{Threat Modeling} discussed in section \ref{section:Threat Modeling}, in particular, in terms of attacker's goal, attacker's knowledge, influence and capability, and attacker's role.

\subsubsection{Root Cause of Adversarial Sampling}\mbox{}\\
One of the main research areas in ML security is adversarial sampling based attacks. \emph{Adversarial sampling}, also known as adversarial input perturbation, intentionally perturbs a small portion of training/test/input data as an attempt to compromise the integrity, availability, or confidentiality of an ML system.

Typically, in the course of an ML system development, the test dataset is drawn from the same distribution as the training dataset. Large sets of the data domain remains unexplored by model learners~\cite{8677311}. In addition, due to some linear model behavior~\cite{Zhang2020,10.1145/2976749.2978392}, the decision boundary is extrapolated to vast regions of high-dimensional subspace that are unpopulated and untrained~\cite{8723317,INDYK20191}. This practice in ML development fails to guarantee model generalization to a different distribution of input data space~\cite{8779838}, and does not account for adversarial samples which often fall outside of the expected input distribution~\cite{7478523}. In fact, adversarial samples are intentionally created by perturbing training/test data or input data into these empty hyper-volumes to compromise model training or mislead the prediction of trained model. Basically, there are three approaches for adversarial sample generation: perturbation on valid samples, transferring adversarial samples across different learner models, and generative adversarial networks (GANs)~\cite{PrithvirajJoseph}.

The design of an ML system should take care of the entire input space. Various technologies  to address these security threats by enhancing model robustness or detecting anomaly input are discussed in section \ref{section:Defense Mechanisms}.

\subsubsection{Poisoning Attack}\mbox{}\\
Machine Learning is vulnerable to attacks at the model training phase (and re-training phase). This type of attacks is called \emph{Poisoning Attack}, which attempts to inject a small fraction of ``poisoned'' samples into training/test dataset in order to modify the statistical characteristics of the dataset, so that the compromised ML model suffers increased rate of misclassified samples at the prediction phase~\cite{BIGGIO2018317,WANG201912}. Poisoning attack is considered a \emph{causative attack} that aims to compromise both the integrity and availability of an ML system. An attacker may launch \emph{error-generic poisoning attacks} that aims to cause an ML system yield as many false outputs as possible so that the ML system becomes unusable to end users (\emph{compromise of availability}), or the attacker may launch \emph{error-specific poisoning attacks} that aims to cause an ML system yield specific outputs as what the attacker desired (\emph{compromise of integrity}), e.g. output an specific incorrect classification.

Typically, an ML training dataset is considered confidential and is well protected from unauthorized access during model training. However, in some cases such as a malware detection system or a spam email filter, in order to adapt to the changing application scenes an ML system might need to continuously re-train its model by taking samples out of the inputs from untrusted sources during its daily operation. A few feasible scenarios of model retraining, including adaptive facial recognition system, malware classification, and spam detection, were discussed in~\cite{Liu2018}. Model retraining brings in an attack surface to adversaries to poison the data for re-training by feeding the operational ML system with adversarial inputs. In summary, poisoning attack can be launched as a \emph{white-box} attack during the initial ML model training phase while it is limited to the attacker's capacity to access and manipulate the training dataset, and it can also be launched as a \emph{black-box} attack during the model re-training phase that the attacker has a viable attack surface but lacks the essential knowledge about the trained model to facilitate the attacks.

\paragraph{\textbf{Adversarial Sampling Algorithms}}\mbox{}\\
There are several types of poisoning attacks algorithms against supervised ML models in terms of the way adversarial samples are generated~\cite{WANG201912,Gardiner:2016:SML:2988524.3003816,Liu2018,shafahi2018poison}. Some of them are described below.

\begin{enumerate}
\item[$\bullet$] \emph{Label-flipping Attack}. This type of attack introduces label noise into training data by flipping the labels, e.g. reverse the label of an amount of legitimate email samples in the training data as spam, and vice versa. Label-flipping Attack can compromise \emph{integrity} or \emph{availability} of an ML system, and is a type of \emph{causative attacks}. Several flipping algorithms are used to generate adversarial samples~\cite{WANG201912,Xiao2012,Biggio2011}, including random label flipping (RLF), nearest-prior label flipping (NPLF), farthest-prior label flipping (FPLF), farthest-rotation label flipping (FRLF), and adversarial label flipping (ALF).

\item[$\bullet$] \emph{Clean Label Attack}. 
This type of attack on neural nets are targeted. They aim to misclassify one specific test instance. For example, they manipulate a face recognition engine to change the identity of one specific person, or manipulate a spam filter to allow/deny a specific email. Clean label attacks do not require control over the labeling function; the poisoned training data appear to be labeled correctly according to an expert observer. This makes the attack not only difficult to detect, but opens the door for attackers to succeed without any inside access to the data collection/labeling process. For example, an adversary could place poisoned images online and wait for them to be scraped by a bot that collects data from the web.

\item[$\bullet$] \emph{Gradient Descent Attack}. Gradient descent-based poisoning attack is a type of \emph{causative}, \emph{availability-compromised} attacks that inserts adversarial samples in to training dataset to \emph{maximize} the impact on an ML system performance, e.g. by reducing the performance to the level that the ML system is unusable. Gradient descent attack is commonly used with label-flipping attack by first flipping the label of a benign training data and then moving it to maximize learner's objective function by leveraging the gradient descent function. Gradient descent-based attack is computationally demanding. In~\cite{BIGGIO2018317}, the researchers reported an optimized algorithm called \emph{back-gradient poisoning} that has much better performance, in terms of reducing the classification accuracy, than random label flipping methods.

\end{enumerate}

\paragraph{\textbf{Backdoor and Trojaning Attack}}\mbox{}\\

With the increasing application of deep neural networks and transfer learning, a specific type of poisoning attacks arose called \emph{Backdoor and Trojaning attack}~\cite{BIGGIO2018317,Liu2018-1,Kissner2019,yan2021dehib}. Backdoor and Trojaning attack can be launched by creating pre-trained network models that include \textit{backdoors} inside. The manipulated models are then released publicly. In the case the models are adopted by innocent users to integrate in their ML system, the attacker can activate the backdoors using specific inputs to mislead the ML system to yield the outputs that they desired. Backdoors can also be inserted by poisoning the training dataset. Backdoor and Trojaning attack is considered a type of \emph{causative}, \emph{integrity-compromising attacks}. 

Recent research shows certain AI systems in neural networks, such as facial recognition \cite{wang2019neural}, self-driving \cite{nassi2020phantom}, are more vulnerable to backdoor attacks. Backdoor attack models mostly use labeled training data in the supervised environment \cite{saha2020hidden}, however, there are some works in the semi-supervised or unsupervised learning environment using unlabelled data \cite{yan2021dehib}. Semi-supervised learning models can be classified into two main categories: Pseudo-label and perturbation based learning. In pseudo-label based approach, pseudo-labels are generated for the unlabeled data to be trained using various methods such as using temporal context: a running average or moving average of past model predictions \cite{tarvainen2017mean}, label propagation in the feature space \cite{iscen2019label}, data augmentations \cite{xie2019unsupervised}. The perturbation-based approaches \cite{miyato2018virtual} use perturbed pictures on training data to provide predictions that are congruent with those of the original images. When compared to pseudo-label-based approaches, these methods perform even worse and require more computation to approximate the Jacobian matrix.

\paragraph{\textbf{Attacks against Federated Learning}}\mbox{}\\
 In federated leaning, participants submit model changes to construct federated models. To protect the privacy of the training data, the aggregator cannot see how these updates are created. This makes model poisoning a threat to federated learning \cite{bagdasaryan2020backdoor}. While standard poisoning attacks target just the training data, model poisoning attacks take use of the fact that federated learning allows malicious participants to exert direct control over the joint model, enabling considerably more effective attacks than training-only poisoning attack. To defend against adversarial manipulations, byzantine-robust models have been proposed. However, researchers~\cite{fang2020local} were able to attack those models, suggesting more research is necessary to improve defense techniques.

\paragraph{\textbf{Attacks against Recommender Systems}}\mbox{}\\
Studies have shown that recommender systems may be deceived to promote a desired item to as many users as feasible. Attacks like these fall into three categories: data poisoning attacks (a.k.a. shilling attacks), profile pollution attacks (a.k.a. phishing attacks), and item/attribute interference attacks (a.k.a privacy attacks)  \cite{huang2021data,xing2013take,fang2018poisoning, fang2020influence}. Attackers inject fake users into a recommender system, changing the recommendation lists. To execute a poisoning attack, the attacker first registers a number of fake users on a web service associated with the recommender system. Each fake user creates custom ratings for a subset of items. This fake data will be included in the target recommender system's training dataset, poisoning the training process. Profile pollution attacks \cite{xing2013take,yang2017fake} use cross-site request forgery to pollute a user's profile (e.g., historical behavior), for example, attacks to recommender systems in web services like Amazon, YouTube etc. Privacy attacks infer the items that a target user has previously rated, for example, products purchased on Amazon, music listened to on Last, and books read on Library. Their attacks are based primarily on the premise that a collaborative filtering recommender system makes recommendations based on user behavior in the past. As a result, the recommendations made by a recommender system incorporate data about the users' previous behavior. In particular, attribute interference attacks in which attackers are provided with a set of users whose ratings and attributes are used to train a machine learning classifier that takes a user's rating behavior as input and predicts the user's attributes. The attacker then uses this classifier to infer user attributes that have not been disclosed. Cambridge Analytica's use of Facebook users' rating behavior (e.g., page likes) to infer users' attributes, which is then used to deliver targeted advertisements to users, is a notable example of a real-world attribute inference attack.

\paragraph{\textbf{Attacks against Unsupervised Learning}}\mbox{}\\
We have found very few works that analyze the effect of adversarial attacks against unsupervised machine learning algorithms. Nonetheless, unsupervised learning models are also vulnerable to adversarial attacks. In~\cite{rubinstein2009antidote}, the authors devised a technique called ``Boiling Frog'' to slowly poison PCA-based unsupervised anomaly detectors. Since online anomaly detectors retrain the models periodically to capture the current pattern of data, the authors showed that it is possible to boost the false negative rate by slowly adding useless data. In~\cite{kloft2010online}, the authors showed that online centroid anomaly detectors are not secure when an attacker controls 5-15\% of all network traffic. In~\cite{chhabra2019strong}, the authors proposed a black-box adversarial attack against four popular clustering algorithms. They also carried out a study on transferability of cross-technique adversarial attack.      

\subsubsection{Evasion Attack}\mbox{}\\
The attacks against an ML system during the prediction phase is called \emph{Evasion Attack}, which can be \emph{error-generic} or \emph{error-specific}~\cite{BIGGIO2018317}. The attack is a type of \emph{exploratory}, \emph{integrity-compromising} attacks that aims to evade the trained model by elaborately manipulating input samples.

\paragraph{\textbf{Adversarial Sampling Algorithms}}\mbox{}\\
Gradient-based attacks apply a gradient descent function to find a state for adversarial samples that mislead an ML model to yield incorrect result. Gradient-based algorithms are widely used to attack differentiable learning algorithms such as DNNs and the SVMs with differentiable kernels. For non-differentiable learning algorithms, such as decision trees and random forests, they are still vulnerable to gradient-based attacks as an attacker can leverage a differentiable surrogate learner~\cite{BIGGIO2018317}. The following adversarial sampling algorithms, designed specifically against DNNs used in the area of computer vision/image classification, are summarized in~\cite{WANG201912}:
\begin{enumerate}
\item[$\bullet$] L-BFGS (Limited-memory Broyden–Fletcher–Goldfarb–Shanno) algorithm. L-BFGS is an optimization algorithm that uses a limited amount of computer memory to approximate BFGS algorithm to find imperceptive perturbations to images that can mislead trained DL models to yield misclassifications~\cite{Szegedy2014}.

\item[$\bullet$] FGSM (Fast Gradient Sign Method) algorithm. FGSM is an efficient adversarial sample generation method that creates samples by appending noise to the original image along the gradient directions~\cite{Tramer2018}.

\item[$\bullet$] UAP (Universal Attack Approach) algorithm. Both L-BFGS and FGSM generate adversarial samples for one single image at a time. The adversarial perturbations cannot be transferred from one image to another. UAP algorithm was developed to generate ``universal'' adversarial perturbations applicable to the images with the same distribution as the images used to generate the perturbations~\cite{Moosavi-Dezfooli2017}. UAP has been validated on ResNet and it was claimed to be effective on various neural networks. 

\item[$\bullet$] UPSET and ANGR algorithms. Both UPSET (Universal Perturbations for Steering to Exact Targets) and ANGR (Antagonistic Network for Generating Rogue) algorithms~\cite{Sarkar2017} are \emph{black-box} attack methods. They have been reported to achieve favorable performance against DL models trained on CIFAR-10 and MNIST datasets.

\item[$\bullet$] C\&W algorithm. C\&W attack, introduced by \emph{C}arlini and \emph{W}agner, is a powerful adversarial sample generation algorithm that achieves better performance in terms of computation speed~\cite{Carlini2017}. It has been reported to achieve impressive results on distilled and undistilled DNN models.

\item[$\bullet$] DeepFool algorithm. DeepFool~\cite{moosavi2016deepfool} algorithm finds the closest distance from original input to the decision boundary of adversarial samples based on an iterative linearization of the classifier. DeepFool algorithm provides an efficient and accurate way to evaluate the robustness of classifiers and to enhance their performance by proper fine-tuning.

\item[$\bullet$] JSMA algorithm. The Jacobian-based Saliency Map (JSMA) algorithm was designed by Papernot {\em et al.}~\cite{papernot2016limitations} to efficiently generate adversarial samples based on computing forward derivatives. JSMA computes the Jacobian matrix of a given sample $x$ to identify input features of $x$ that made the most significant changes to the output classification. While JSMA adds smaller perturbations in a smaller portion of features than FGSM, it is much slower due to its significant computational cost. 
\end{enumerate}

\paragraph{\textbf{Transferable Adversarial Samples}}\mbox{}\\
Researchers observed that the adversarial samples generated for a trained model by some algorithms, such as C\&W, can be transferred (effective) against another trained model~\cite{BIGGIO2018317,WANG201912,Liu2018,VMB18}. This enables an attacker who does not have perfect knowledge of an ML system to be able to conduct black-box attack against the ML system. The attacker can develop a \emph{surrogate model} by training it using surrogate training data, generate and test adversarial samples against the surrogate model, and then apply the adversarial samples against the victim ML system.

\paragraph{\textbf{Mimicry Attack}}\mbox{}\\
\emph{Mimicry attack} is a type of evasion attacks that was used to attack traditional ML models. With the emergence of DNN/DL, mimicry attack is used together with the gradient-based methods to attack neural networks~\cite{Gardiner:2016:SML:2988524.3003816,Wagner2002}. Mimicry attack attempts to modify the features of adversarial samples such that the adversarial samples mislead a trained model to classify them as benign inputs. For example, mimicry attack has been used to bypass an ML-based IDS by hiding the traces of system calls that actually carried out malicious activity. Mimicry attack can also be used to attack unsupervised ML algorithms, e.g. clustering algorithms, by effectively reducing the distance between the adversarial samples and benign inputs.

\begin{table*}[htb]
	\centering
	\caption{Attack Vectors}
	\label{tabsec4:attackvectors}
\resizebox{\columnwidth}{!}{%
\begin{tabular}{|m{1.3cm}|m{2.0cm}|m{2.0cm}|m{2.5cm}|m{1.5cm}|m{1.6cm}|m{1.6cm}|m{1.6cm}|}
		\hline
		\multicolumn{1}{|m{1.3cm}|}{\textbf{Attack Vector}} & \multicolumn{1}{m{2.0cm}|}{\textbf{Citation}} & \multicolumn{1}{m{2.0cm}|}{\textbf{Attacking Method}}                & \multicolumn{1}{m{2.5cm}|}{\textbf{Adversarial Algorithm}} & \multicolumn{1}{m{1.5cm}|}{\textbf{ML Pipeline}} & \multicolumn{1}{m{1.6cm}|}{\textbf{Attacker's Influence}} & \multicolumn{1}{m{1.6cm}|}{\textbf{Security Violation}} & \multicolumn{1}{m{1.6cm}|}{\textbf{Attacker's Knowledge}}                       \\ \hline
		
	\multirow{4}{\linewidth}{Poisoning Attack}   & \cite{WANG201912,Xiao2012,Biggio2011}                                                                 & Label-flipping                        & RLF, NPLF, FPLF, FRLF, ALF                             & \multirow{4}{\linewidth}{Model (re-) training}  & \multirow{4}{*}{Causative}                & Integrity \& Availability               & \multirow{4}{*}{White-box}                \\ \cline{2-4} \cline{7-7}
	& \cite{BIGGIO2018317,Gardiner:2016:SML:2988524.3003816}                                                & Gradient Descent                    & Back-gradient descent                                  &                                  &                                           & Availability                            &                                           \\ \cline{2-4} \cline{7-7}
	& \cite{BIGGIO2018317,Liu2018-1,Kissner2019}                                                            & Backdoor \& Trojaning                 & ---                                                    &                                  &                                           & Integrity                               &                                           \\ \cline{2-4} \cline{7-7}
	& \cite{rubinstein2009antidote,kloft2010online}                                                            &      ---           & Boiling frog, Greedy optimal attack                                                &                                  &                                           & Integrity \& Availability                              &                                           \\ \hline
	\multirow{4}{\linewidth}{Evasion Attack}     & \multirow{2}{\linewidth}{\cite{WANG201912,Szegedy2014,Tramer2018,Moosavi-Dezfooli2017,Sarkar2017,Carlini2017,moosavi2016deepfool,papernot2016limitations}} & \multirow{2}{*}{\parbox{5cm}{Gradient\\ descent}}     & L-BFGS, FGSM, UAP, C\&W, JSMA, DeepFool                                & \multirow{4}{*}{Prediction}      & \multirow{4}{*}{Exploratory}              & \multirow{4}{*}{Integrity}              & White-box                                 \\ \cline{4-4} \cline{8-8} 
	&                                                                                                       &                                       & UPSET AND ANGR                                         &                                  &                                           &                                         & Black-box                                 \\ \cline{2-4} \cline{8-8} 
	& \cite{BIGGIO2018317,WANG201912,Liu2018,VMB18}                                                         & Transferable samples                  & C\&W                                                   &                                  &                                           &                                         & White-box                                 \\ \cline{2-4} \cline{8-8} 
	& \cite{Gardiner:2016:SML:2988524.3003816,Wagner2002}                                                   & Mimicry                               & ---                                                    &                                  &                                           &                                         & Black-box                                 \\ \hline
\end{tabular}%
}
\end{table*}

\subsection{Defense Mechanisms}
\label{section:Defense Mechanisms}

People often assume ML models are trained, tested and deployed in a benign setting. This assumption is not always valid. An ML system should be designed with the consideration of adversarial settings in mind in which capable adversaries can access and elaborately manipulate the training/test data and/or input data to compromise the integrity, availability, or privacy of the ML system. These risks should be analyzed and addressed when an ML system is designed \cite{8779838}. In section \ref{section:Attack Vectors}, various attack vectors were discussed. In this section, the corresponding countermeasures against the attacks at the model training phase and the prediction phase will be discussed. These countermeasures are summarized in Table~\ref{tabsec4:defense}. 

\subsubsection{Model Enhancement}\mbox{}\\
The \emph{Model Enhancement} mechanism attempts to improve the robustness of the trained models during the model training phase by leveraging various methods ~\cite{WANG201912} discussed in detail below. 

\paragraph{\textbf{Adversarial Training}}\mbox{}\\
\emph{Adversarial Training} is essentially a robust generalization method~\cite{Shaham2018,Goodfellow2015}. It adds and mixes adversarial samples into the original training dataset to enhance the model robustness against the attacks using these adversarial samples. This method is not adaptive to different types of adversarial sampling attacks~\cite{WANG201912}, which means the model has to be trained on relevant adversarial samples in order to resist a particular type of adversarial attacks. 

Adversarial training is a heuristic approach that has no formal guarantees on convergence and robustness properties~\cite{BIGGIO2018317,PrithvirajJoseph}. Some researchers leveraged the game theory computational framework to enhance model robustness through adversarial training~\cite{PrithvirajJoseph}, in which both a Learner (such as a classifier) and an Adversary can be utilized to learn a prediction mechanism from each of the other party. From the Learner's perspective, the adversarial training techniques can be used as a defense method at the model training phase to make the trained model more robust against the adversarial attacks. GAN-based methods have been used to construct robust DL models against FGSM-based attacks~\cite{WANG201912,Wagner2002,Lee2017}. The authors reported that the trained models can successfully classify original and contaminated images, and even rectify perturbed images.

A more efficient adversarial training method called \emph{robust optimization} which formulates adversarial training as a \emph{MiniMax} problem~\cite{BIGGIO2018317}: the inner problem maximizes the training loss by manipulating training data under bounded, worst-case perturbation, while the outer problem trains the learner to minimize the corresponding worst-case training loss. The robust optimization aims to smooth out the decision boundary to make it less sensitive to worst-case input manipulation.

\paragraph{\textbf{Data Compression}}\mbox{}\\
Researchers found out that various data compression methods can counter adversarial sampling attacks against image classifiers~\cite{WANG201912}. For example, JPG compression and JPEG compression can mitigate FGSM-based adversarial sampling attacks by removing high frequency signal components inside square blocks of an image~\cite{Das2017,Dziugaite2016,Guo2018}. These compression defending methods, however, may lead to the decrease in the classifier's accuracy when the compression rate is set high.

\paragraph{\textbf{Foveation-based method}}\mbox{}\\
The foveation mechanism, which selects a region of the image to apply a Convolutional Neural Network (CNN) while discarding information from the other regions, can be used to mitigate adversarial attacks against image classifiers~\cite{WANG201912,Luo2015}. Researchers observed that the CNN model which has been enforced by the foveation mechanism is robust to scale and transformation changes over the images. This method has not yet been validated against more powerful attacks.

\paragraph{\textbf{Gradient Masking}}\mbox{}\\
The gradient masking method enhances ML model robustness by modifying the gradients of input data, loss or activation function~\cite{WANG201912}. The method can defend against L-BFGS and FGSM based adversarial attacks by penalizing the gradient of loss function of neural networks~\cite{Lyu2016} or minimizing loss function of neural networks over adversarial samples~\cite{Shaham2018} when model parameters are updated. The method can also defend against C\&W attacks by adding noise to a neural network’s logit output against the low distortion attacks~\cite{Nguyen2018}. Researchers also found that gradient regularization is helpful to improve model robustness as it penalizes the variation degree of training data during model training~\cite{BIGGIO2018317,WANG201912,Ros2018}.

\paragraph{\textbf{Defensive Distillation}}\mbox{}\\
Distillation is a technique originally used to reduce DNN dimensionality. Papernot {\em et al.} devised a variant of the method, called defensive distillation, to enhance the model generalizability and robustness that can significantly reduce the effectiveness of adversarial perturbations against DNNs~\cite{WANG201912,Papernot2016}. Defensive distillation extracts the knowledge from a trained DNN model and then uses the knowledge to re-train the model to enhance the resistance to adversarial attacks.

\paragraph{\textbf{DeepCloak}}\mbox{}\\
The DeepCloak method identifies and then removes unnecessary features in a CNN model, which can enhance the robustness of the model as the method limits attackers' capacity to generate adversarial samples~\cite{WANG201912,Gao2019}. By applying DeepCloak, a masking layer is inserted between the convolutional layer(s) and the fully connected layer(s) of a DNN model. The deepcloak layer is then trained using original and adversarial image pairs. Since the most prominent features have the dominant weights, the prominent features can be removed by masking the dominant weights for the deepcloak layer.

\paragraph{\textbf{Certified Defense}}\mbox{}\\
Certified defense is a defense mechanism that recently attracts great interest among machine learning security researchers. Different from the empirical defense methods discussed above, certified defense can provide provable robustness to certain kinds of, often norm-bounded, adversarial perturbations. 

~\cite{Raghunathan2018} is the first work claimed to demonstrate a method that can train a simple (two-layer networks) model and certify the robustness of the model against adversarial attacks during the inference. Lecuyer {\em et al.} ~\cite{Lecuyer2019} propose the first certified defense, called \emph{PixelDP}, which scales to large neural networks and datasets and can be applied to arbitrary DNN model types. In this paper, the authors formally establish a novel connection between model robustness and differential privacy mechanisms, and leverage the DP-robustness connection to train robust models and evaluate the robustness of the models against norm-bounded adversarial examples at inference time. \emph{PixelDP} enhances the robustness of DNN models by two means; (i) adding a DP noise layer in the network which randomizes the network’s computation to enforce DP bounds on its predictions with small, norm-bounded changes in the input; and (ii) replacing the original scoring function with a DP scoring function using Monte Carlo methods that can estimate the robustness for each individual prediction. In ~\cite{Cohen2019}, Cohen {\em et al.} leverage the randomized smoothing method that transforms any arbitrary base classifier \emph{f}, including large-scale artificial neural networks, into a new ``smoothed classifier" \emph{g} that is certifiably robust in $\ell_2$ norm with radius \emph{R}, where for any input \emph{x}, \emph{g(x)} returns the most probable prediction by \emph{f} of random Gaussian corruptions of \emph{x}. A set of Monte Carlo algorithms are used to evaluate \emph{g(x)} and certifying the robustness of \emph{g} around \emph{x}. The base classifier is trained on noised data with Gaussian data augmentation that was proposed in~\cite{Lecuyer2019}. One major contribution of the paper is to prove that the $\ell_2$ robustness guarantee is \emph{tight} in that for an arbitrary base classifier it is impossible to certify an $\ell_2$ ball with radius larger than \emph{R}. While the work in ~\cite{Lecuyer2019} and ~\cite{Cohen2019} focuses on certified defense for top-\emph{1} prediction, the authors in ~\cite{Jia2019} advance it by deriving the tight certified robustness under $\ell_2$ norm for top-\emph{k} prediction.

A key idea of the certified defense method is to leverage a \emph{majority vote mechanism} to interpret the outcome of trained models to predict the label for each input. Jia {\em et al.} found out that such majority vote mechanism is inherent in some machine learning models, such as \emph{k} nearest neighbors (kNN) and radius nearest neighbors (rNN) ~\cite{Jia2020} and bootstrap aggregating (bagging) ensemble learning method ~\cite{Jia2020a}. These models are certifiably robust against data poisoning attacks when the poisoned training examples is bounded.

\subsubsection{External Defense Layer}\mbox{}\\
The \emph{External Defense Layer} mechanism adds an extra layer and attempts to process input data before or after they are sent to the trained models during the prediction phase. The defense mechanisms that fall into the category of External Defense Layer include \emph{Input Monitoring} and \emph{Input Transformation}. Input Monitoring tries to detect and filter adversarial input samples e.g. using an anomaly detection system, while Input Transformation tries to sanitize suspicious input samples e.g. those are sufficiently far from the training data in feature space~\cite{8779838,BIGGIO2018317,8677311}. In~\cite{li2019desvig}, the authors tried to combine these two techniques. They proposed a decentralized system which takes both the input and the label from the deep learning models and uses a conditional generative adversarial network (CGAN) to detect adversarial examples and also suggest the correct input/label back to the original DL model. 

\paragraph{\textbf{Input Monitoring}}\mbox{}\\
\emph{Input Monitoring} against anomaly input data is a defense mechanism that can be applied at both the model training phase and model prediction phase \cite{8779838}.

\emph{Feature Squeezing} is an input monitoring method that is used to test input images. It uses two feature squeezing techniques: reducing the color bit depth of each pixel and spatial smoothing~\cite{Xu2018}. It then compares the model classification accuracy on the original images and the squeezed images~\cite{WANG201912}. If there exists substantial difference between the accuracy, the input images are considered as adversarial samples.

In~\cite{8677311}, Darvish Rouani {\em et al.} presented their ML defense mechanism - \emph{Adaptive ML model assurance}. The researchers developed an external module called \emph{modular robust redundancy (MRR)} to thwart potential adversarial attacks and keep the trained ML model intact so that the performance of the ML system is not impacted. 

Carrara {\em et al.}~\cite{Carrara2019} proposed a \emph{scoring} approach to detecting adversarial samples for kNN (k-nearest neighbors) learning algorithm used for image classification. The method defines an authenticity confidence score based on kNN similarity searching among the training images, and analyzes the activations of the neurons in hidden layers (deep features) to detect adversarial inputs. The deep features are assumed to be more robust to adversarial samples for two reasons: (i) the adversarial sample generation algorithms are meant to fool final classification but not deep features; and (ii) generated adversarial samples look similar to authentic ones for humans, and deep features have shown impressive performance in visual similarity/differentiation related tasks. It was reported that the method can filter out many adversarial samples while retaining most of the correctly classified authentic images.

\paragraph{\textbf{Input Transformation}}\mbox{}\\
Machine Learning models may include extraneous information in its learned, hidden representations which are not relevant to the ML learning tasks \cite{8779838}. An attacker can conduct attacks against the ML system by taking advantage of these extraneous information. \emph{Input Transformation} mechanism can be used to defend against this type of attacks by attenuating or discarding this extraneous variation.

\emph{Perturbation Rectifying Network (PRN)} is a universal perturbation defense framework to effectively defend DNNs against UAP attacks~\cite{WANG201912,Akhtar2017}. A PRN is learned from real and synthetic image-agnostic perturbations. A separate perturbation detector is trained on the \emph{Discrete Cosine Transform} of the input-output difference of the PRN. If a perturbation is detected, the output of the PRN is used for label prediction instead of the actual input. Therefore, the PRN can process input images and detect possible perturbations, and then rectify the images before sending them to the classifier.

\subsubsection{Defense against Attacks During Training Phase}\mbox{}\\
During the model training phase, ML is vulnerable to Poisoning Attack that attempts to insert adversarial samples into the training dataset or modify the statistical characteristics of existing dataset to compromise the trained model. To defeat these attacks, there are two common countermeasures: \emph{Data Sanitization} and \emph{Robust Learning}~\cite{BIGGIO2018317,WANG201912,Liu2018}:

\paragraph{\textbf{Data Sanitization}}\mbox{}\\
\emph{Data Sanitization} is a defense technique that tests and identifies abnormal input samples and then removes them from the training dataset. In order to impact the ML learner in a negative way, adversarial samples have to exhibit different statistical characteristics. Therefore, data sanitization technologies, which are able to detect anomaly training data by analyzing discrepancies in the statistical characteristics, can be used to filter out adversarial samples. Cretu {\em et al.} proposed a sanitization scheme that is reported to significantly improve the quality of unlabeled training data by making “attack-free” dataset~\cite{Cretu2008}. Nelson {\em et al.} proposed a \emph{Reject On Negative Impact} (RONI) defense method that has been used to protect several ML-based spam filters~\cite{Liu2018,Nelson2009}. RONI tests the impact of each email on training and discards the messages that have a large negative impact.

\paragraph{\textbf{Robust Learning}}\mbox{}\\
\emph{Robust learning} is a defense technique that optimizes learning algorithms so that models are learned based on robust statistics that are intrinsically less sensitive to outlying training samples~\cite{BIGGIO2018317}. Robust learning hardens ML learners by improving the generalization capability. In~\cite{Globerson2006}, the author introduced a new algorithm for avoiding single feature over-weighting so that the trained classifiers are optimally resilient to deletion of features. The method was illustrated in the application scenarios of spam filtering and handwritten digit recognition.

\begin{table*}[htb]
\centering
\caption{Defense Mechanisms}
\label{tabsec4:defense}
\resizebox{\columnwidth}{!}{%
\begin{tabular}{|m{1.5cm}|m{2.5cm}|m{2.5cm}|m{2.5cm}|m{1.5cm}|m{5.0cm}|}
\hline
\textbf{Defense Mechanism}              & \textbf{Citation} & \textbf{Defense Method} & \textbf{Algorithm or Framework}                   & \textbf{ML Pipeline}            & \textbf{Notes} \\ \hline
\multirow{6}{1.5cm}{Model Enhance\-ment}      & \cite{BIGGIO2018317,PrithvirajJoseph,WANG201912,Wagner2002,Shaham2018,Goodfellow2015,Lee2017}                   & Adversarial Training       & Game theory, GAN, Robust Optimization             & \multirow{6}{*}{Prediction}     & A heuristic approach that has no formal guarantees on convergence and robustness properties                                                                                                              \\ \cline{2-4} \cline{6-6} 
                                        & \cite{WANG201912,Das2017,Dziugaite2016,Guo2018}                   & Data Compression           & JPG, JPEG &                                 & Mitigate FGSM-based attacks against image classifiers                                           \\ \cline{2-4} \cline{6-6} 
                                        & \cite{WANG201912,Luo2015}                  & Foveation-based            & Object Crop MP, Saliency Crop MP, 10 Crop MP, 3 Crop MP                                  &                                 & Mitigate scale and transformation perturbations against DNN image classifier                                  \\ \cline{2-4} \cline{6-6} 
                                        & \cite{BIGGIO2018317,WANG201912,Shaham2018,Lyu2016,Nguyen2018,Ros2018}                   & Gradient Masking           & ---                                  &                                 & Mitigate L-BFGS, FGSM, or C\&W-based adversarial attacks against neural network-based image classifiers \\ \cline{2-4} \cline{6-6} 
                                        & \cite{WANG201912,Papernot2016}                   & Defense Distillation       & ---                                  &                                 & Decrease the dimensionality of trained DNN models                                                             \\ \cline{2-4} \cline{6-6} 
                                        & \cite{WANG201912,Gao2019}                   & DeepCloak                  & DeepCloak                                  &                                 & Remove prominent features of convolutional neural network-based image classifiers                             \\ \hline
\multirow{2}{\linewidth}{External Defense Layer} & \cite{8779838,WANG201912,8677311,Xu2018,Carrara2019}
                   & Input Monitoring           & Feature Squeezing; MRR               & \multirow{2}{*}{Prediction}     & Defend image classifiers                                                                                      \\ \cline{2-4} \cline{6-6} 
                                        & \cite{8779838,WANG201912,Akhtar2017}                   & Input Transformation       & Perturbation Rectifying Network       &                                 & Defend image classifiers                                                                                      \\ \hline
\multirow{2}{\linewidth}{Secure Training Data}   & \cite{BIGGIO2018317,WANG201912,Liu2018,Cretu2008,Nelson2009}                   & Data Sanitization          & RONI                                  & \multirow{2}{\linewidth}{Model Training} & Leverage data sanitization technologies                                                                       \\ \cline{2-4} \cline{6-6} 
                                        & \cite{BIGGIO2018317,Globerson2006}                   & Robust Learning            & Feature Deletion                                  &                                 & Improve the generalization capability of the learning algorithms                                              \\ \hline
\end{tabular}%
}
\end{table*}

\section{Privacy-Preserving Machine Learning}
\label{section:ML Privacy}

In the previous section, we have presented various attacks that compromise the \emph{integrity} and \emph{availability} of ML systems, and corresponding multi-stage defense mechanisms. We now turn our attention to the various ways in which \emph{confidentiality} or \emph{privacy} may be compromised in an ML pipeline and some protection measures. Table~\ref{tab: ML Privacy} summarizes our findings.

\subsection{Types of Privacy Breaches}
\emph{Statistical disclosure control} states that the model should reveal no more about the input to which it is applied than would have been known about this input without applying the model. A related notion of privacy appears in~\cite{fredrikson2014privacy}: a privacy breach occurs if an adversary can use the model’s output to infer the values of sensitive attributes used as input to the model. However, it is not always possible to prevent this kind of privacy breach if the model is based on statistical facts about the population. For example, the model may breach privacy not of the people whose data was used to create the model, but also of other people from the same population, even those whose data was not used and whose identities may not even be known. Valid models generalize accurate predictions on inputs that were not part of their training datasets. That is, the creator of a generalizable model cannot do anything for the privacy protection because the correlations on which the model is based or the inferences that these correlations enable exist for the entire population, regardless of the training sample or the model creation.

Machine Learning \emph{training data} usually contains a large amount of private and confidential information. Meantime, \emph{trained models} including model hyperparameters are also considered sensitive information since adversaries can take advantage of the knowledge of these information to launch more powerful, white-box or gray-box based attacks. There are four common types of attacks during the ML training and prediction phases to steal these information~\cite{8677282,tramer2016stealing}.

\subsubsection{Reconstruction Attack}\mbox{}\\
\emph{Reconstruction attacks} reconstruct raw, private training data by using the knowledge of model feature vectors~\cite{8677282}. Reconstruction attacks are \emph{white-box attacks} which require access to an ML model's parameters such as model feature vectors. In the cases where the feature vectors are not removed from the trained model when the model is deployed in production, or for some learning algorithms such as SVM or kNNs, where those feature vectors are stored with the model, reconstruction attack is possible. Examples of this type of attacks include fingerprint reconstruction and mobile device touch gesture reconstruction. A countermeasure against reconstruction attacks is to avoid using ML models that store explicit feature vectors (e.g. SVM).

\subsubsection{Model Inversion Attack}\mbox{}\\
\emph{Model inversion attacks}~\cite{8677282,fredrikson2015model} utilize the responses for inputs sent to a trained ML model in an attempt to create feature vectors that resemble those used in model training. If the responses include confidence information of model prediction, the attack can produce an \emph{average of confidence} that represents a certain class. Typically, the model inversion attack does not infer whether a sample was in the training dataset or not. However, in cases where a certain class represents an individual, e.g. in the application scenario of face recognition, the individual's privacy might be breached.

Model inversion attack can be launched together with reconstruction attack to further breach ML privacy. To resist this type of attack, researchers advised limiting the information included in the responses from model prediction. For example, in the case of classifier models, classification algorithms should only report rounded confidence values or even just the predicted class labels.

\subsubsection{Membership Inference Attack}\mbox{}\\
\emph{Membership inference attacks} attempt to determine if a sample was a member of the training dataset~\cite{8677282}. Shokri {\em et al.}~\cite{shokri2017membership} quantitatively explore how machine learning models leak information about the individual data records. Given a data record and black-box access to a model, they try to determine if the record was in the model’s training dataset. To perform membership inference against a target model, they trained their own model to recognize differences in the target model’s predictions on the inputs that it trained on versus the inputs that it did not. 

Overfitting is an important reason why machine learning models leak information about their training datasets. Regularization techniques such as dropout can help defeat overfitting and also strengthen privacy guarantees in neural networks. Regularization is also used for objective perturbation in differentially private machine learning. 

\subsubsection{Model Extraction}\mbox{}\\
In \emph{model extraction attacks}, an adversary with black-box access and no prior knowledge of an ML model’s parameters or training data, can query an ML model to obtain predictions on input feature vectors. The adversary’s goal is to extract an equivalent or near-equivalent ML model including model parameters~\cite{tramer2016stealing} and even hyperparameters~\cite{8418595}, and to duplicate the functionality of the model. Unlike in classical learning settings, ML-as-a-Service offerings may accept partial feature vectors as inputs and include confidence values with predictions. Given these practices, the authors showed simple, efficient equation-solving model extraction attacks that use non-adaptive, random queries to extract target ML models with near-perfect fidelity for popular model classes including logistic regression, neural networks, and decision trees. They demonstrated these attacks against the online services of BigML and Amazon Machine Learning and showed that the natural countermeasure of omitting confidence values from model outputs still admits potentially harmful model extraction attacks.

\subsubsection{Classic Privacy Considerations}\mbox{}\\
Training deep learning models is a computationally and data intensive task. As it was discussed in section~\ref{section:Threat Modeling}, typically the computation party (cloud computing servers, machine learning as a service providers, etc.) is a different entity from the input party (the owner or contributor of the training data). In this case the model training task is \emph{delegated} to the computation party. How to prevent privacy breach of the training data during the stages of data collection, data transition, and data storage becomes a classic privacy-preserving problem. Furthermore, in the setting of Machine Learning as a Service (MLaaS), trained models are only available through a cloud service to end users. From the user's perspective, there are privacy concerns when they supply samples to the service for prediction. These concerns on data privacy are primarily addressed by using various cryptographic mechanisms which will be discussed in detail in the next subsection.

\subsection{Privacy-Preserving Measures}
Privacy-Preserving ML (PPML) technologies protect against some of the privacy breaches described above and enable \emph{collaborative} learning, in which the input party and the computation party are distinct entities. There are two main techniques to protect ML privacy~\cite{WANG201912,8677282,Liu2018}: \emph{Perturbation Mechanisms} including differential privacy methods and dimensionality reduction methods, and \emph{Cryptographic Mechanisms} including fully homomorphic encryption,  secure  multiparty computation, functional encryption, and crypto-oriented model architectures.

\subsubsection{Differential Privacy}\mbox{}\\
\emph{Differential Privacy} (DP)~\cite{dwork_calibrating_2006,dwork_differential_2006} is a method to quantify and control the risk to one's privacy when personal information is included in a dataset. To achieve differentially private models, the applicable defenses follow the paradigm of \emph{security-by-obscurity}~\cite{BIGGIO2018317}, which obscure training data by adding noise at various points  to protect privacy. Examples of this technique include input perturbation (noised added to input data), algorithm perturbation (noise added to intermediate values in iterative learning algorithms), output perturbation (noise added to trained models), and objective perturbation (noise added to objective function for learning algorithms). Differentially private models are resistant to membership inference attacks.

Researchers have proposed some specific model architectures to satisfy DP. 
Randomized Aggregative Privacy-Preserving Ordinal Response (RAPPOR) proposed by Erlingsson {\em et al.}~\cite{10.1145/2660267.2660348} is a method that employs randomized response mechanisms to achieve differential privacy in the context of  crowdsourced datasets. Private Aggregation of Teacher Ensembles (PATE) proposed by Papernot {\em et al.}~\cite{papernot2018scalable} protects the privacy of trained models by constructing a Teacher-Student model that prevents adversaries from having direct access to trained Teacher model, as a way to protect the training data and the Teacher model parameters. 

Li {\em et al.}~\cite{LI2018341} proposed a differentially private scheme called privacy-preserving machine learning under multiple keys (PMLM) which supports multiple data providers to securely share encrypted datasets with a cloud server for model training. The PMLM scheme uses public-key encryption with a double decryption algorithm (DD-PKE) to transform the encrypted data into a randomized dataset without information leakage.

\subsubsection{Dimensionality Reduction}\mbox{}\\
\emph{Dimensionality Reduction} (DR) methods project continuous and high-dimensional training data to a lower dimensional hyperplane to prevent adversaries from reconstructing original data or inferring sensitive information. This method offer resistance against reconstruction attacks.

Hamm {\em et al.}~\cite{Jihun17} proposed a DR-based defense method applied at the model training phase. The method uses a MiniMax filter combined with a differentially-private mechanism to transform continuous and high-dimensional raw features to dimensionality-reduced representations. This preserves the information on target tasks, but sensitive attributes of the data are removed which makes it difficult for an adversary to accurately infer such sensitive attributes from the filtered output. The min-diff-max filter\, defined formally in Equation~(\ref{eq:min-diff-max}), is designed to achieve an optimal utility-privacy trade-off in terms of prediction accuracy and expected privacy risks. 
\begin{equation} \label{eq:min-diff-max}
  \min_u\Phi(u) = \min_u[\max_v-f_{\mathrm{priv}}(u,v) -\rho \max_w -f_{\mathrm{util}}(u,w)]
\end{equation}

This formula is to simultaneously resolve two problems of maximizing privacy and minimizing disutility, where $u$ and $v$ are the filter parameters and $w$ is the inference model parameters. The author demonstrated in the paper that the proposed method achieves similar or higher target task accuracy and significantly lower inference accuracy based on several experiments on real-world tasks.

\subsubsection{Adversarially Crafted Noise}\mbox{}\\
\label{section:adv-privacy}
Privacy attacks are sometimes carried out by the attacker training a machine learning model for a specific objective. For example, in membership inference attacks the adversary trains a model to infer whether a data point was part of the training set. A possible defense against such attacks is to add noise to the parts of the pipeline the attacker targets that is adversarially crafted with the goal of triggering a misclassification on the adversary's model. Since the adversary's model is unknown, the adversarial examples are built using another model for the same task. Transferability of adversarial examples ensures that the resulting noise will remain effective against the attacker's model. This method can be used to defend against membership inference attacks~\cite{10.1145/3319535.3363201}, but also against attacks that aim to recover private attributes from publicly available information~\cite{217523} or against identification~\cite{chandrasekaran2020face}.



\subsubsection{Homomorphic Encryption}\mbox{}\\
\label{sec:homoencphil}
\emph{Homomorphic encryption} (HE)~\cite{gentry09} is a set of cryptographic
algorithms that allow computations to be
performed on the underlying plaintext by acting solely on the encrypted
ciphertext. More formally, $\mathsf{Enc}$ is a Fully\footnote{``Fully'' because it can compute any boolean circuit homomorphically.} Homomorphic Encryption (FHE) if it is equipped with operations ``$\oplus$'' and ``$\otimes$'' such that $\mathsf{Enc}(a)\oplus \mathsf{Enc}(b)=\mathsf{Enc}(a+b)$ and $\mathsf{Enc}(a)\otimes \mathsf{Enc}(b)=\mathsf{Enc}(a\times b)$.  Several challenges arise when appying FHE to machine learning:
\begin{itemize}
\item FHE algorithms are defined for polynomials over finite rings
  (e.g. the integers modulo $n$ with addition and multiplication forms
  a ring) whereas ML operates over floating point
  numbers.
\item It adds significant computing overhead to any operation. For
  most of the existing FHE schemes, the time complexity of \emph{each}
  elementary operation ($+$ and $\times$) grows with the
  multiplicative depth of the \emph{total} circuit to evaluate. This
  depth must be known in advance in order to select the appropriate
  parameters for the encryption scheme.
\item It is restricted to computing polynomial functions. While
  polynomials can in theory compute any function as they can emulate
  boolean circuits, the multiplicative depth of the resulting function
  would make such implementations impractical. Hence a common approach
  in the works cited in this section is to use polynomial
  approximation of more complex functions.
\end{itemize}
Homomorphic encryption has evident application to privacy-preserving prediction and training where the input party's data is hidden from the computation party: the input party encrypts its input and the computation party performs operations homomorphically on the ciphertext.

Other variants of homomorphic encryption offer a tradeoff by homomorphically evaluating restricted classes of boolean circuits, but generally having more efficient implementations. \emph{Partially homomorphic encryption} only allows the homomorphic evaluation of $+$ or $\times$, but not both. \emph{Leveled homomorphic encryption} can evaluate boolean circuits of a fixed pre-determined depth. These more efficient modes are advantageous for ML applications, but restrict the number of non-linearities that may appear in the model architecture. For example, partially HE can only homomorphically compute linear functions and leveled HE can homomorphically compute more complex function such as neural network feed forward, but only up to a certain number of layers.

\paragraph{\textbf{Prediction Phase}}\mbox{}\\
Graepel {\em et al.} \cite{graepel_ml_2013} built algorithms for
training and classification composed exclusively of low-degree
polynomials. Prediction accuracy is negatively impacted and training is restricted to simple models.
Gilad-Bachrach {\em et al.} \cite{gilad-bachrach_cryptonets:_2016} perform privacy-preserving prediction for neural networks by encoding real numbers into polynomial
rings and using low-degree polynomial approximation of commong non-linear functions (e.g. ReLU). 
Chabanne {\em et al.}~\cite{chabanne_privacy-preserving_2017} use the observation that
polynomial approximations of activation functions (e.g. ReLU) are more accurate
on small values centered around $0$. They exploit this insight by introducing
batch normalization to the FHE prediction framework of~\cite{gilad-bachrach_cryptonets:_2016} to achieve better prediction accuracy on a DNN.
By changing the way the input vectors are represented and by using a more
advanced HE scheme which implements certain types of rotations homomorphically,
Brutzkus {\em et al.}~\cite{brutzkus_low_2019} drastically reduce the latency
and memory usage of~\cite{gilad-bachrach_cryptonets:_2016}. They 
evaluate DNNs homomorphically using transfer learning. 

\paragraph{\textbf{Training and Feature Selection}}\mbox{}\\
The limitations of HE are evident in early attempts to apply it to
privacy-preserving training of even a simple model.
In~\cite{nikolaenko_privacy-preserving_2013}, the input party delegates training of a ridge regression model using HE for linear operations and a trusted third party (TTP) that 
performs secure multiparty computation (see Section~\ref{sec:mpcphil}) with the computation party for non-linear operations (e.g. max). 
Zhang {\em et al.} \cite{zhang_privacy_2016} proposed a protocol that adds rounds of interaction
between the input and computation parties where, between each step of the
backpropagation algorithm, the computation party sends back to the input party the encrypted
result of the backpropagation step, the input party decrypts it, updates the model
parameters and sends back the encryption of the updated parameters.

Other components of a machine learning pipeline besides training and prediction
are receiving attention. Masters {\em et al.}~\cite{masters_towards_2020} use
HE for the feature selection and prediction phases of a
ML pipeline also using low-degree polynomial approximations of common machine learning
functions.

End-to-end training of large models on encrypted data is something that is still out of
reach with the current state-of-the-art due to the overhead of HE on
already computationally intensive operations. One of the downsides is that the
whole training pipeline has to be automated as it is not possible for data
scientists to inspect the training data. One of the challenges facing broader
adoption of privacy-preserving prediction is that the state-of-the-art techniques in terms
of efficiency can be expensive as these techniques require special model architectures that need to be planned in
advance of training and retraining large models.

\subsubsection{Secure Multiparty Computation}\mbox{}\\
\label{sec:mpcphil}
Existing technology in homomorphic encryption is very computationally
expensive, especially if evaluating complex functions
homomorphically. Another solution researchers have turned to is \emph{secure
multiparty computation} (MPC)~\cite{lindell_secure_2020}. In MPC, two
or more distrustful parties want to compute a joint function of their
private inputs without revealing anything more about their inputs than
the function output.


Bost {\em et al.} \cite{bost_machine_2015} show how MPC can speed up privacy-preserving prediction compared to HE if we allow interaction. They conceive new MPC
protocols optimized for a core set of operations performed during classification
tasks such as comparison, argmax and dot product. Makri {\em et al.}~\cite{makri_epic:_2019} consider the setting of privacy-preserving prediction that employs multiple intermediate parties (called MPC
servers) of which at least one must be honest to ensure privacy. 

A natural setting where MPC can be used is in distributed modes of learning
where many participants train a joint neural network from their respective
datasets. Danner {\em et
al.}~\cite{danner_fully_2015} propose an efficient MPC solution to gradient aggregation for \emph{gossip learning} where many parties contribute to the training of a
joint model without using a centralized entity or broadcast communications.
Another popular form of distributed learning is \emph{federated
 learning}~\cite{pmlr-v54-mcmahan17a} where gradient aggregation is performed
at a centralized server that sends back the updated model after each aggregation
step. Bonawitz {\em et al.}~\cite{bonawitz_practical_2017} apply an MPC layer over the
federated learning framework to ensure data privacy by a secure gradient
aggregation procedure where the server only learns the aggregated gradients, but
not the individual gradients. 


The use of MPC in PPML is usually more efficient than FHE alone, but requires synchronicity of all parties, incurs communication costs and delays, and sometimes requires additional actors~\cite{makri_epic:_2019}. Furthermore,
most of the papers cited in this section provide a weak notion of security where the
adversary is assumed to act semi-honestly -- they try to gain information while
following the protocol. In general, adversaries may act maliciously --
arbitrarily deviating from the protocol -- to gain more information. An important research topic is to strengthen the security guarantees provided by the schemes cited here.

\subsubsection{Functional Encryption}\mbox{}\\
\label{sec:funcencphil}
\emph{Functional encryption} (FE) is a primitive akin to homomorphic encryption with
the distinction that the holder of an encryption of $x$ may learn the value
$f(x)$ through the use of a decryption key associated with $f$, while leaking
only the value $f(x)$ and nothing more about $x$. Functional encryption can be
used when the input party wants the computation party to act upon the value $f(x)$,
for example in spam classification.

Sans {\em et al.}~\cite{sans_reading_2018} show that while the above use case is beyond
existing FE technologies, some applications of FE to privacy-preserving
classification are currently feasible. They demonstrate a classifier for MNIST
digits based on a one hidden layer neural network with quadratic activation functions where
a decryption key allows the server to learn the classification outcome. Their
private classification framework is based on their own FE scheme for quadratic
functions which improves efficiency compared to previous results.
The idea of using FE for delegated ML purposes was pushed further
by Marc {\em et al.}~\cite{marc_privacy-enhanced_2019} who apply it to multiple privacy-preserving
ML tasks and provide an open-source functional encryption
library with implementation of common FE primitives used in ML
such as inner product, the square function and attribute-based encryption.

Functional encryption is useful when the party evaluating the ML model on private data needs to take action based on the outcome, which would require the data holder to decrypt and send back the result using FHE, or when the outcome of the prediction should not be known to the data holder. However, efficient implementations of FE are only known for some restricted classes of functions such as the inner product or quadratic functions~\cite{Tilen_FE}.



\subsubsection{Crypto-Oriented Model Architectures}\mbox{}\\
\label{sec:hybridsphil}
It should now be clear that special care needs to be taken when running machine
learning alongside homomorphic encryption or secure multiparty computation,
otherwise the time and communication complexity can spiral out of control.
Beyond polynomial approximations of non-linearities in neural networks, recent
results approach the problem of privacy-preserving prediction or training as a
neural network design challenge.

Bourse {\em et al.}~\cite{bourse_fast_2018} propose discretized neural networks where model weights are integers and the
chosen activation function is the sign function $z\mapsto z/|z|$. They show that
pre-trained conventional neural networks can be converted into discretized
neural networks and how to boost efficiency through scale invariance in the complexity of HE
operations.
Mishra {\em et al.} \cite{244032} propose an hybrid
HE + MPC approach to privately evaluate neural network architectures that uses
both ReLUs (executed using interactive MPC) and quadratic approximations
(executed non-interactively using HE). The original contribution of this work
is a planning procedure that uses techniques similar to hyperparameter search to
find which activations of a neural architecture to replace with quadratic
approximations and which to keep as ReLUs.
Shafran {\em et al.} \cite{shafran2019crypto} propose a new kind of neural network
architecture based on \emph{partial activation layers} where activation
functions (e.g. ReLU) are only applied to a fraction of the neurons in that
layer, the rest acting only as linear operations. Through experiments, they
demonstrate that partial activation layers can achieve a good tradeoff between
accuracy and efficiency.

\begin{table*}[htb]
 \centering
\caption{ML privacy \& Countermeasure }
\label{tab: ML Privacy}
\resizebox{\columnwidth}{!}{%
\begin{tabular}{|l|l|l|l|}
\hline
 \multicolumn{1}{|c|}{\textbf{Citation}}&
  \multicolumn{1}{c|}{\textbf{Privacy Breaches}} &
  \multicolumn{1}{c|}{\textbf{Countermeasures}} &
  \multicolumn{1}{c|}{\textbf{ML Pipeline stage}} \\ \hline

\cite{shokri2017membership} &
  \begin{tabular}[c]{@{}l@{}}Membership Inference\end{tabular} &
  \begin{tabular}[c]{@{}l@{}}Regularization techniques\\ Differential privacy\end{tabular} &
  \begin{tabular}[c]{@{}l@{}}Prediction Phase\end{tabular} \\ \hline

\cite{fredrikson2015model} &  

\begin{tabular}[c]{@{}l@{}}Model Inversion\\Reconstruction\\Deblurring\end{tabular} &
  \begin{tabular}[c]{@{}l@{}}Privacy aware decision Tree\end{tabular} &
  \begin{tabular}[c]{@{}l@{}}Training Phase\end{tabular} \\ \hline

\cite{fredrikson2014privacy} &  

\begin{tabular}[c]{@{}l@{}}Model Extraction\\Path finding\end{tabular} &
\begin{tabular}[c]{@{}l@{}}Differential privacy\\Rounding confidence\\Ensemble method\end{tabular} &
\begin{tabular}[c]{@{}l@{}}Training Phase\\Prediction phase\end{tabular} \\ \hline

\cite{tramer2016stealing} & 

\begin{tabular}[c]{@{}l@{}}Model Inversion\\\end{tabular} &
\begin{tabular}[c]{@{}l@{}}Differential privacy\end{tabular} &
\begin{tabular}[c]{@{}l@{}}Training Phase\end{tabular} \\ \hline


  \cite{graepel_ml_2013, gilad-bachrach_cryptonets:_2016,bourse_fast_2018,chabanne_privacy-preserving_2017,brutzkus_low_2019,masters_towards_2020} &\multirow{6}{*}{Privacy-Preserving Training/Prediction}&FHE \& polynomial approximation&
    \begin{tabular}{@{}l}
      Training Phase\\Prediction Phase
    \end{tabular}
 \\\cline{1-1}\cline{3-4}

    \cite{bost_machine_2015,makri_epic:_2019} && \multirow{2}{*}{MPC}&
      Prediction phase
  \\\cline{1-1}\cline{4-4}

    \cite{bonawitz_practical_2017} &&
& \multirow{2}{*}{Distributed training} 
  \\\cline{1-1}\cline{3-3}

    \cite{danner_fully_2015} &&
\begin{tabular}{@{}l}
  Secret sharing \& additive HE
\end{tabular}
&    
  \\\cline{1-1}\cline{3-4}
  
      \cite{244032,shafran2019crypto,nikolaenko_privacy-preserving_2013,zhang_privacy_2016} &&
\begin{tabular}{@{}l}
  FHE + MPC
\end{tabular}
&  \multirow{2}{*}{Prediction phase}
  \\\cline{1-1}\cline{3-3}

    \cite{sans_reading_2018,marc_privacy-enhanced_2019} &&
\begin{tabular}{@{}l}
  Functional encryption
\end{tabular}
&
  \\\hline
  
\end{tabular}%
}
\end{table*}

\section{User Trust}
\label{section:User Trust}
While there is a lot of focus on the technical aspects of cyber secure systems, less attention is paid to the interactions between the users and the systems, including research into how end users deal with cyber security attacks. Jajali {\em et al.}~\cite{jalali2019health} conducted a systematic review on cyber security of health care information systems and found that the majority of articles were technology-focused. There is an increasing demand for studies on user trust in machine learning systems. Our literature review investigated the issues surrounding user behaviours and trust of ML-based systems, how user trust plays an important role in user acceptance of ML systems, the security implication as users react to cyber security attacks, and the factors that have impacts on user trust. We further studied design principles and best practices to increase user trust. Our investigation is focused on human factors of ML systems in health care and autonomous vehicles.

\subsection{User trust and adoption of machine learning based systems}
A common theme of articles on user acceptance and machine learning boosted healthcare systems is that of trust; researchers identified trust as an important factor in user attitudes towards IoT based healthcare~\cite{alraja2019effect,yusif2016older,hengstler2016applied,al2016progress}. In a  review of articles on barriers to older people’s adoption of assistive technologies, researchers found that the top concern when adopting assistive technologies is related to privacy, followed by issues of trust~\cite{yusif2016older}. Likewise Jaschinski \& Allouch found that privacy and intrusiveness were the most important barriers to acceptance – both in the interviews and evaluation phases of their study~\cite{jaschinski2019listening}. In healthcare it is important to ensure that users do not feel like they are under surveillance, as it decreases the user's willingness to accept such technologies; is easier for user to accept technologies if they feel in control. As a result, there is a need in design to keep user engaged~\cite{hengstler2016applied}. Systems perceived as intrusive can lead to lower levels of user acceptance – a fact that many researchers overlooked~\cite{cavoukian2010remote}.

Autonomous vehicles is an emerging market where research and commercialization is growing steadily. The rise of artificial intelligence (AI) based automated decision systems brings the need/right/obligation (due to regulations) for user explanations of outcomes to increase trust of the system's decision. Beyond that, the lack of trust in automation as a result of ML tools results in an issue with user adoption~\cite{o2019question}. Challenges in the adoption of autonomous vehicles includes issues of trust and ethical implications, both which pose serious threats to user acceptance of the technology~\cite{adnan2018trust}. Privacy and cyber security risks of autonomous vehicles are critical to building consumer trust. Informational privacy, which includes the protection of data against misuse, building consumer trust and safeguarding against surveillance~\cite{lim2018autonomous}, is needed to receive the benefits of personal data while controlling the risks.

\subsection{User reaction to security attack}
While partial automation in vehicles has been in use for decades, high and fully autonomous vehicles represents a relatively new level of technology. Assets in level 4 (high autonomous) and 5 (full automation) vehicles~\cite{EuropeanUnionAgencyforCybersecurity.2019} are numerous, likewise potential threats to autonomous vehicles are vast. There is a large body of research on the potential for cyber security attacks on this technology, but not a lot of research on how users react to such attacks. While there has not been any example of real world cyber-attacks on autonomous vehicles, there have been plenty of experimental white hat attacks such as remote attacks on autonomous vehicles to enable, disable or manipulate the systems; attacks against various in-vehicle sensors e.g. GPS spoofing; and physical attacks such as drawing white lines/circle around the vehicle to trap it~\cite{EuropeanUnionAgencyforCybersecurity.2019}.

There is yet to be any studies on user trust in autonomous vehicles to help users react appropriately and timely to cyber security attacks, which is problematic for the average user especially when using new technology and experiencing cognitive overload~\cite{linkov2019human}. This could lead to issues related to the security of the system if a user ignores or overrides any warnings or alerts generated by the system. Parkinson {\em et al.} in 2017 identified 14 cyber threats facing autonomous and connected vehicles. Amongst the list they identified several knowledge gaps related to reactions to errors at run-time. In particular they noted a lack of research on how the vehicle and the driver would react to detection of a cyber-attack – how would the vehicle behave during a suspected attacks and how information could be given to the user in order to make any necessary decisions~\cite{parkinson2017cyber}. Likewise, Linkiov {\em et al.} state that there is a current need to study how people might behave during a cyber-attack on their autonomous vehicle. They note that cyber-attacks generally cause increased stress on the user. However, there is no current research on how users react during an attack on a vehicle they are riding in – there needs to be research on how cyber security issues can be communicated to the user in order to elicit appropriate reactions~\cite{linkov2019human}.

\subsection{User trust: design principles and best practices}
The aforementioned research on the interplay of user trust with system adoption and cyber security leads researchers to propose further studies on user-centered-design principles and best practices.

\subsubsection{Increase user visibility}\mbox{}\\
User visibility has an effect on user trust – there is a need for greater efforts if the product has a higher level of visibility, and therefore there is a need for proactive communication~\cite{hengstler2016applied}. One key issue with data sharing and privacy of health care data is that of control. In their study on IoT based healthcare, Alraja {\em et al.} noticed that there was an increase in user trust when they believed they had control over what data people could access on the system. Their trust levels also increased when providers ensured them that their personal data would be protected~\cite{alraja2019effect}.

There is a need to understand not just the technological tools needed to ensure cyber secure healthcare systems but also what the end user wants these systems to look like, how much control they require over such systems and how they interact with such systems. Trust and trade-offs are linked in ML healthcare systems, whereby users will not consider trade-offs between giving up personal information for usability without trusting that their personal information is safe.

\subsubsection{Safety protocol design}\mbox{}\\
There is a lot of literature on attacking autonomous vehicle (from white hat hackers) but an unexplored area is how safety protocol measures could prevent such attacks. Current test drivers are well educated and trained on the protocol once control is given to the user; however those users with lower technical ability and who are unsure about what the safety implications are might have issues with the change in protocol. Also attention needs to be paid to how the driver should be alerted to issues in order to act quickly and safely~\cite{parkinson2017cyber}.

\subsubsection{User trust measurement}\mbox{}\\
There are many limitations in studying trust and user acceptance of autonomous vehicles and healthcare systems. The main limitation is the newness of the technology and limited means in which to test real life uses of the technology. Current studies only examine user intent to use through surveys~\cite{woldeamanuel2018perceived} or simulations~\cite{molnar2018understanding}. More attention needs to be paid to how trust is measured as currently there are no appropriate scales to measure trust in autonomous vehicle technology.

\subsubsection{Informed consent over the disclosure of data}\mbox{}\\
Data gathered during or after real world usage of the technology would be valuable. An issue of note is the discussion over the amount of personal data automated vehicles will generate, and the ownership and security protocols around this. Much of the literature is focused on the need for robust anonymization as well as strong encryption~\cite{parkinson2017cyber}, but there are also issues here which could lead into discussions of informed consent over the disclosure of data, and what types of data are necessary (i.e. location data gathered from smartphones compared to user preference data gathered from web browsers) as well as how to disclose when that personal data has been compromised.

\section{Empirical Robustness Assessment}
\label{section:Assessment and Metrics}
It is well recognized that an objective and comprehensive assessment, which covers a set of aspects of the quality of a system including system robustness and human factors, is crucial to building trustworthy systems~\cite{Cho2016}. A trained machine learning model should be measured by the model performance on prediction accuracy and equally important by the capability to resist adversarial attacks~\cite{7478523}. Independent and standard assessment methodologies and metrics for ML model resilience should be devised to support trustworthy ML system development. 

\subsection{Quantitative Analysis}
\emph{Quantitative analysis} is a key tool to assess robustness of ML algorithms against adversarial attacks~\cite{Laskov2009}, in terms of attacker's constraints, strategy of attacking optimization, and adversarial impacts. A realistic assessment of ML security risk is a ``reasonable-case" analysis which is based on reasonable assumptions on attacker's capacity, resource and constraints. Such assumptions, e.g. the fraction of the training data that can be controlled or the content of network traffic can be manipulated by adversaries, may vary among ML systems. Threat modeling discussed in section~\ref{section:Threat Modeling} can be leveraged to systematically identify these assumptions.

Researchers have devised various attacking methods, as discussed in section~\ref{section:ML Offensive and Defensive Technologies} and \ref{section:ML Privacy}, to exploit the weaknesses and vulnerabilities in ML algorithms. The efficiency of the attacks (and defense mechanisms, vice versa) typically are measured by the level of negative impacts on an ML system performance. For binary classifiers, the commonly used performance metrics, including \emph{Accuracy}, \emph{Precision}, \emph{Recall}, and \emph{F1 score}, are derived from four distinctive classification outcomes: true positives (TP), true negatives (TN), false positives (FP), and false negatives (FN)~\cite{Seliya2009}:
\begin{enumerate}
\item[$\bullet$] \emph{Accuracy} = (TP + TN)/(TP + FP + FN + TN): represents the ratio of correctly predicted samples to the total samples;
\item[$\bullet$] \emph{Precision} = TP/(TP + FP): represents the ratio of correctly predicted positive samples to the total predicted positive samples;
\item[$\bullet$] \emph{Recall} = TP/(TP + FN): represents the ratio of correctly predicted positive samples to the all positive samples; and
\item[$\bullet$] \emph{F1 score} = 2$\times$(Recall $\times$ Precision)/(Recall + Precision): represents the weighted average of Precision and Recall that takes into account both false positives and false negatives. 
\end{enumerate}

\begin{figure}[htbp]
\centerline{\includegraphics[width=0.5\textwidth]{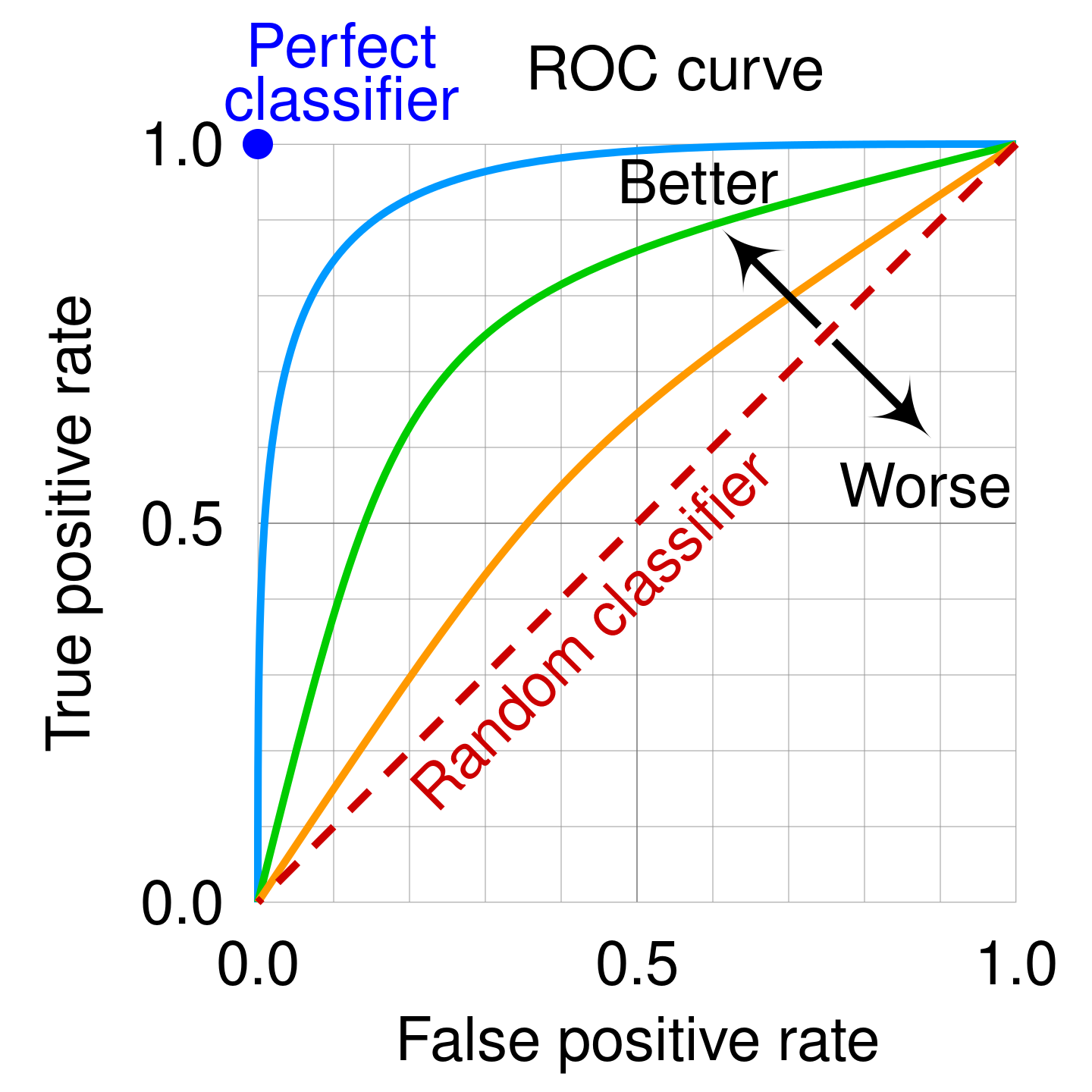}}
\caption{The ROC space for a "better" and "worse" classifier.}
\label{fig:ROC}
\end{figure}

ROC (Receiver Operating Characteristic) curve, as illustrated in Figure~\ref{fig:ROC}~\footnote{By cmglee, MartinThoma - Roc-draft-xkcd-style.svg, CC BY-SA 4.0, https://commons.wikimedia.org/w/index.php?curid=109730045}, demonstrates the performance of a binary classifier by plotting the TPR (true positive rate) against the FPR (false positive rate) at various threshold settings. ROC is often used with AUC (Area Under the Curve) by researchers to assess and compare binary classifiers. 

These metrics have been widely used by researchers for various ML algorithms and applications to demonstrate performance decrease when an ML system suffers from various adversarial attacks. Depending on the factors such as security violation (e.g. integrity vs. availability), attack vectors (e.g. evasion attack vs. poisoning attack), and application scenarios, the suitable metrics may vary. For instance, in~\cite{Dunn2020} the metrics Accuracy, Precision, False Positive, and True Positive were used to measure the negative impacts of the poisoning attacks to four ML models' integrity, including gradient boosted machines, random forests, naive Bayes statistical classifiers, and feed forward deep learning models, in the application of IoT environments; in~\cite{Sadeghzadeh2020}, the evaluation was focused on the Recall metric to study the robustness of a Deep Learning-based network traffic classifier by applying several Universal Adversarial Perturbation based attacks against various traffic types including chat, email, file transfer, streaming, torrent, and VoIP; in~\cite{Biggio2014}, several metrics including ROC, AUC, Genuine Acceptance Rate (GAR) and False Acceptance Rate (FAR) were proposed to evaluate pattern classifiers used in adversarial applications like biometric authentication, network intrusion detection, and spam filtering. 

\subsection{Assessment Methodologies}
It is a very challenging task to perform ML system security evaluation, especially when assessing the efficiency of defense mechanisms. Studies have reported that many defense mechanisms that were claimed efficient have been found out either less efficient (lower robust test accuracy) or even broken when enhanced, diversified attacks were used ~\cite{Carlini2019,Croce2020,Goodfellow2018}, which led to wrong impression of ML system robustness.

Researchers have proposed novel methods to address this challenge. Biggio {\em et al.} proposed an empirical security evaluation framework that can be applied to different classifiers, learning algorithms, and classification tasks~\cite{Biggio2014}. The framework, which consists of an adversary model to define any attack scenario, a corresponding data distribution model, and a method for generating training and testing sets used for empirical performance evaluation, provides a quantitative and general-purpose basis for classifier security evaluation.

Goodfellow {\em et al.} introduced the concept of ``attack bundling"~\cite{Goodfellow2018}. While various attack algorithms can be used to generate adversarial samples, the attack bundling is devised to measure the true worst-case error rate with the consideration of threat modeling by choosing the strongest adversarial sample among the ones generated using all the available attack algorithms for each clean example. While the approach of attack bundling is considered suitable for white-box attacks, the authors also described the way to apply it in the black-box setting. Attack bundling may help to alleviate the problem of ML robustness overestimation.

In~\cite{Croce2020}, Croce and Hein presented their work on ML robustness evaluation. The authors designed a new gradient-based scheme called Auto-PGD which remedies the standard Projected Gradient Descent (PGD) attacks, the most popular adversarial algorithm, by (i) automatically adjusting the hyperparameter ``step size" and (ii) using an alternative loss function. Auto-PGD is then combined with the white-box FAB-attack and the black-box Square Attack to form an ensemble of complementary attacks called AutoAttack to increase attack diversities. AutoAttack has been studied in a large-scale evaluation on over 50 classifiers from 35 papers that claimed the models are robust. The reported result demonstrates the efficiency of the tool with the majority of the tests yielding lower robustness than the ones claimed in the original papers and several identified broken defenses.

Carlini {\em et al.} in their paper~\cite{Carlini2019} discussed the methodologies and best practices in ML robustness evaluation, including the principles of rigorous evaluation, the common flaws and pitfalls, and the recommendations on evaluation tasks and analysis. The document is claimed by the authors a live document that may help ML security practitioners as a guidance to develop and evaluate defensive ML technologies. 

The fact that lack of suitable robustness metrics hinders widespread adoption of robust ML models in practice~\cite{Gardiner:2016:SML:2988524.3003816} attracted some researchers' interests. 
In~\cite{BIGGIO2018317}, Biggio {\em et al.} proposed assessing and selecting ML learning algorithms and models based on the \emph{security evaluation curve}, which measures the extent to which the performance on prediction accuracy of a trained model drops under attacks of increasing attack strength (for example, the amount of input perturbation for evasion attacks or the number of adversarial samples injected into training data for poisoning attack). The authors further argued that while the metric of minimally-perturbed adversarial samples can be used to analyze the sensitivity of a trained model, maximum-confidence attacks are more suitable for assessing model resilience. That is, using the security evaluation curve to demonstrate that if attack strength is not larger than $\epsilon$, then the model prediction performance should not drop more than $\delta$. 

Katzir {\em et al.} proposed a formal metric called model robustness (MRB) score to evaluate the relative resilience of different ML models, as an attempt to quantify the resilience of various ML classifiers applied to cyber security~\cite{KATZIR2018419}. The method is based on two core concepts - total \emph{attack budget} and \emph{feature manipulation cost} to model an attacker's abilities. The researchers reported that MRB provides a concise and easy-to-use comparison metric to compare the resilience of different classification models trained using different ML learning algorithms against simulated evasion attack and availability attack. MRB method relies on domain experts to estimate feature manipulation costs, which is prone to subjective variation and is considered a limitation of the method.

\section{Robust and Trustworthy Machine Learning System Development}
\label{section:ML Development}
In the previous sections, we presented various technologies that can be used to support robust and trustworthy ML system development. However, our literature review found few studies on how to leverage these technologies from a security engineering perspective, which in general encompasses tools, methods and processes to support system development to protect the system and its data from malicious attacks~\cite{anderson_2008}. In this section, we attempt to push our effort forward above and beyond a survey by exploring how to address this gap of knowledge. Based on the literature we have studied, we developed a metamodel to formalize and generalize the body of knowledge. We choose to use UML, a de facto general-purpose development and modeling language in the field of software engineering, to specify the core concepts, the fundamental entities and their intricate relationships captured in the metamodel. We then further illustrate how to perform ML threat analysis and security design following a systematic process driven by the metamodel.

\subsection{Robust and Trustworthy ML Development: Metamodel}
\label{section:Metamodel}
\begin{figure}[th!]
	\centering
	\includegraphics[width=1\linewidth]{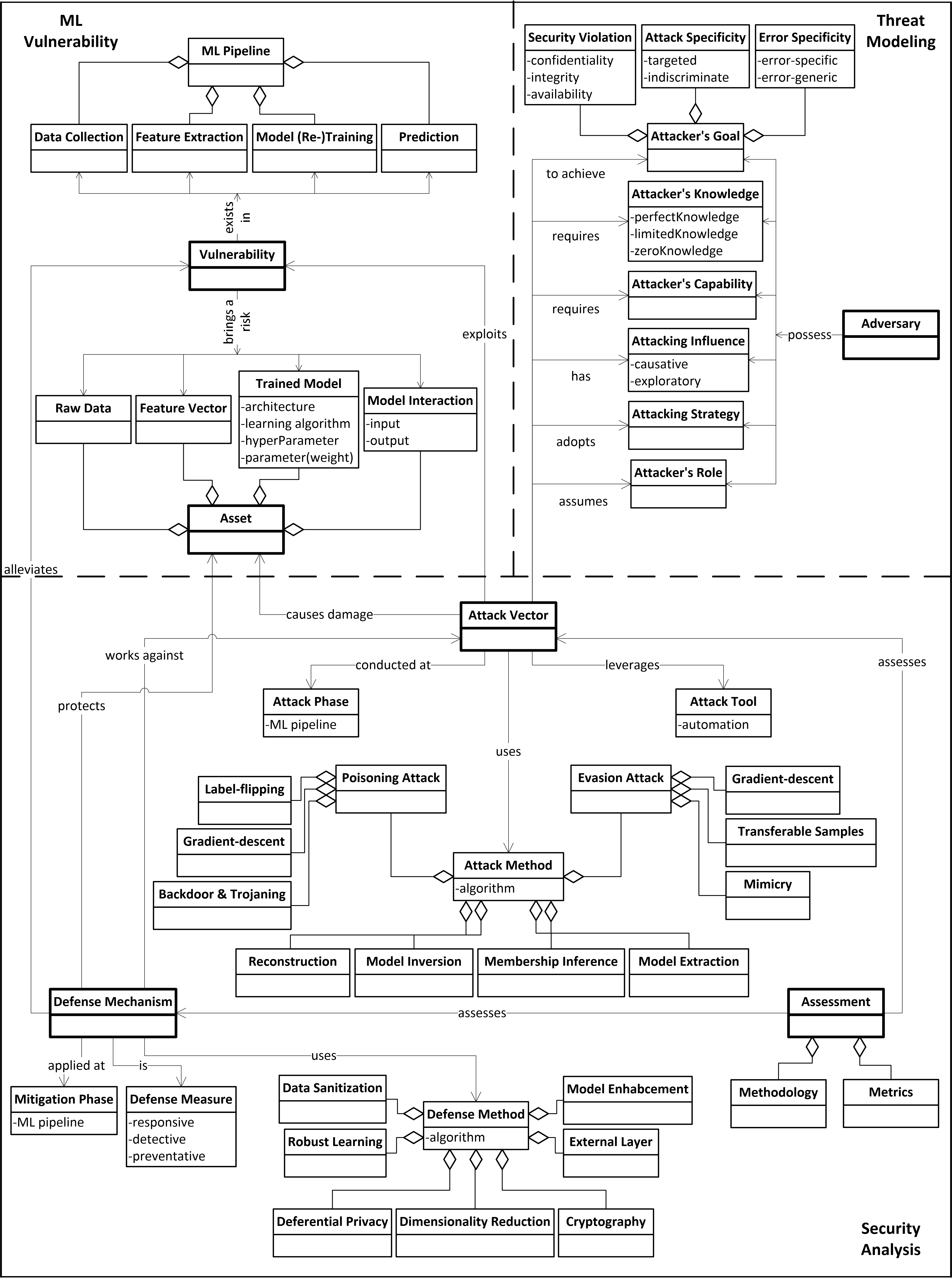}
	\caption{Robust and Trustworthy Machine Learning Development: MetaModel}
	\label{fig:MetaModel}
\end{figure}

Figure \ref{fig:MetaModel} shows the metamodel that captures the entities and their relationships from three different aspects: ML Vulnerability, Threat Modeling, and Security Analysis.

The \emph{ML Vulnerability sub-model} presents assets to be protected across a typical ML pipeline and vulnerabilities that are exploitable and therefore will bring in risks against the assets. By analyzing the data flow along the pipeline, a set of assets has been identified including raw data; feature vectors; information pertinent to model training such as learning algorithms, model architecture and hyperparameters; trained model parameters (weights and biases); and input/output for model prediction. The \emph{Threat Modeling sub-model} captures the adversarial aspects of threat modeling in the context of ML system development that have been discussed in detail in section~\ref{section:Threat Modeling}. It models a comprehensive profile of potential adversaries who pose threats to an ML system. The \emph{Security Analysis sub-model} includes three types of key entities in secure ML development: attack vectors, defense mechanisms, and robustness assessment that have been discussed through section~\ref{section:ML Offensive and Defensive Technologies} to~\ref{section:Assessment and Metrics}. An attack vector represents a path to exploit an ML system by using various attack methods and usually with the help of tools to increase the power of attacking. At each ML pipeline phase there may exist multiple feasible attack vectors. Correspondingly, different defense mechanisms may defeat the attacks and mitigate the risks. Robustness assessment, a critical tool to assure a system's security posture, follows a systematic approach and uses a set of suitable quantitative and qualitative metrics to gauge the performance and efficiency of the applicable attack vectors and adopted defense mechanisms.   

The metamodel provides ML practitioners with an expressive model in a standard and visualized way for robust and trustworthy system development. In the following subsections, we will illustrate how we can perform systematic threat analysis and security design in the context of a generic ML system development by leveraging the metamodel.

\subsection{Threat Modeling}
Threat modeling is a process to define the security objectives for a system, identify potential attackers and their goals and methods, and conclude potential threats that may arise. The \emph{security objectives} is the first artifact we developed during threat modeling. It is set to protect all the assets identified in the ``ML Vulnerability" sub-model including the ML system itself: 
\begin{enumerate}
\item[$\bullet$] protect the authenticity, integrity, and confidentiality of raw data;
\item[$\bullet$] protect the integrity and confidentiality of feature vector;
\item[$\bullet$] protect the integrity and confidentiality of trained model;
\item[$\bullet$] prevent biased model; and 
\item[$\bullet$] prevent system from misuse. 
\end{enumerate}

Based on the ``ML Vulnerability" sub-model, we further developed the \emph{attack surface}, as shown in Figure \ref{fig:AttackSurface} in Appendix~\ref{Appendix}, which identifies all the potential attacking points along the entire pipeline where an ML system may be exposed to adversaries. With the identified attack surface, we developed \emph{attack trees}, as shown in Figure \ref{fig:AttackTrees} in Appendix~\ref{Appendix}, against the defined security objectives. The attack trees are derived from the potential attack vectors specified in the ``Security Analysis" sub-model. The attack trees present at high level an intuitive and visualized view of the threats an ML system may face. It is a very useful tool to further derive attack scenarios that can be used during the system security design phase to help identify appropriate defense mechanisms as well as to validate the system's security compliance during the system implementation phase. 

\subsection{Security Design}
We used the attack trees to effectively identify appropriate defense mechanisms based on the available defense methods specified in the ``Security Analysis" sub-model. Figure \ref{fig:SecurityDesign} in Appendix~\ref{Appendix} shows an example of the security design that adopts various defense methods across the entire ML pipeline to defeat the attacks and protect the system assets. 

It is worth noting that the aforementioned analysis of the threat modeling and security design is based on the context of a generic ML system development. The ``Threat Modeling" sub-model, which identifies various characters of potential adversaries, can be used to depict the adversaries against an ML system. These characters can be used to develop a taxonomy of the attack-defense methods, so that feasible attacks and applicable countermeasures can be easily identified in the context of a specific, concrete ML system.

\section{Conclusion: Open Problems and Future Research Directions}
\label{section:Conclusions}
Machine Learning technologies have been widely adopted in many application areas. Despite the benefits enabled by applying the ML technologies, it is a challenge to ensure that the ML systems are sufficiently robust against security attacks and privacy breaches and users have trust in the systems. Robust and trustworthy ML system development has not yet been widely adopted in industry. From the security engineering perspective, this is due to a number of reasons including the lack of (i) general guidance with key principles and best practices; (ii) efficient ML defensive technologies; (iii) ML robustness assessment methodologies and metrics; and (iv) dedicated tool support. In this article, we summarized our findings on the survey of the state-of-the-art technologies, and our engineering initiative of leveraging the technologies in supporting robust and trustworthy ML system development.

In many research work that we reviewed, the authors emphasized the importance of offensive-defensive ML technologies and proposed the design of robust ML algorithms as a direction to future research. The results of our investigations reinforce that view. In addition, we believe the engineering of ML system development, which is currently at its incipient stage, is a critical cornerstone to ensure ML system's robustness and trustworthiness. In section~\ref{section:ML Development}, we demonstrated what a systematic approach to ML threat modeling and security design may look like, by extending and scaling up the classical process. Our results and their analysis are preliminary since a more thorough treatment would be beyond the scope of this work. Hence, we propose the following two future research directions that we expect could shed light on the research in this area for both industrial and academic practitioners in the near future.

\paragraph{\textbf{Develop a comprehensive ML security metamodel}}\mbox{}\\
Figure \ref{fig:MetaModel} in Appendix~\ref{Appendix} presents a preliminary result of the modeling effort, which is limited to the scope of our survey. A comprehensive metamodel should be developed to represent the body of knowledge considering all of the following aspects: 
\begin{enumerate}
\item[$\bullet$] different machine learning approaches such as supervised learning, unsupervised learning, and reinforcement learning;
\item[$\bullet$] different machine learning architectures including classic, ``shallow" models and deep neural network learning models; and
\item[$\bullet$] all phases in the ML pipeline including data collection, feature extraction, model training, prediction, and model retraining.
\end{enumerate}
In addition, while plenty of research on the ML offensive-defensive technologies are primarily based on various adversarial sampling algorithms, there exist simple and direct attacks across the process of ML system implementation and integration, e.g. last layer attack, GPU overflow attack, tainted open source models and datasets~\cite{Kissner2019,Xiao2018,Stevens2017}, which should be included in the scope of the modeling as well.

\paragraph{\textbf{Develop an ontology for machine learning robustness and trustworthiness}}\mbox{}\\
The ontology of a certain domain is about its terminology, essential concepts in the domain, their classification and taxonomy, their relations, and domain axioms~\cite{dragan_vladan_2009}. The ontology of cyber security was first coined in 2012 by the Software Engineering Institute at Carnegie Mellon University\footnote{https://insights.sei.cmu.edu/insider-threat/2013/03/how-ontologies-can-help-build-a-science-of-cybersecurity.html; visited on 10/03/2020}. Since then  a number of efforts have been made including the recent progress on the ontology for network security~\cite{Silva2017}. Another example of applying ontology in building trustworthy systems is an ontology-based metric framework proposed in~\cite{Cho2016}.

The development of an ontology for ML robustness and trustworthiness includes capturing and representing the basic concepts, key entities and intricate relationships between them in a formal way. UML, which is a suitable language for knowledge representation~\cite{dragan_vladan_2009}, provides a standard way to specify and visualize information system. It can be used to develop structural models that emphasize the organization of a system, and behavioral models that emphasize the dynamics of the system. The metamodel we created is constructed using class diagram - a type of structural model. There are many other types of UML models available to extend our metamodel to further represent the body of knowledge in the domain. A proper and coherent ML ontology can facilitate the ML community to communicate and exchange profound knowledge, develop and share innovative ideas, and open the door to developing tools that support ML system development, as well as can guide ML practitioners to follow a more systematic and efficient development process.

\bibliographystyle{elsarticle-num} 
\bibliography{pulei.bib,heather.bib,crypto.bib,shahrear.bib,mamun.bib,scott.bib}

\pagebreak
\appendix
\section{Supplementary Material}
\label{Appendix}

\begin{figure}[htbp]
\centerline{\includegraphics[width=0.9\paperwidth]{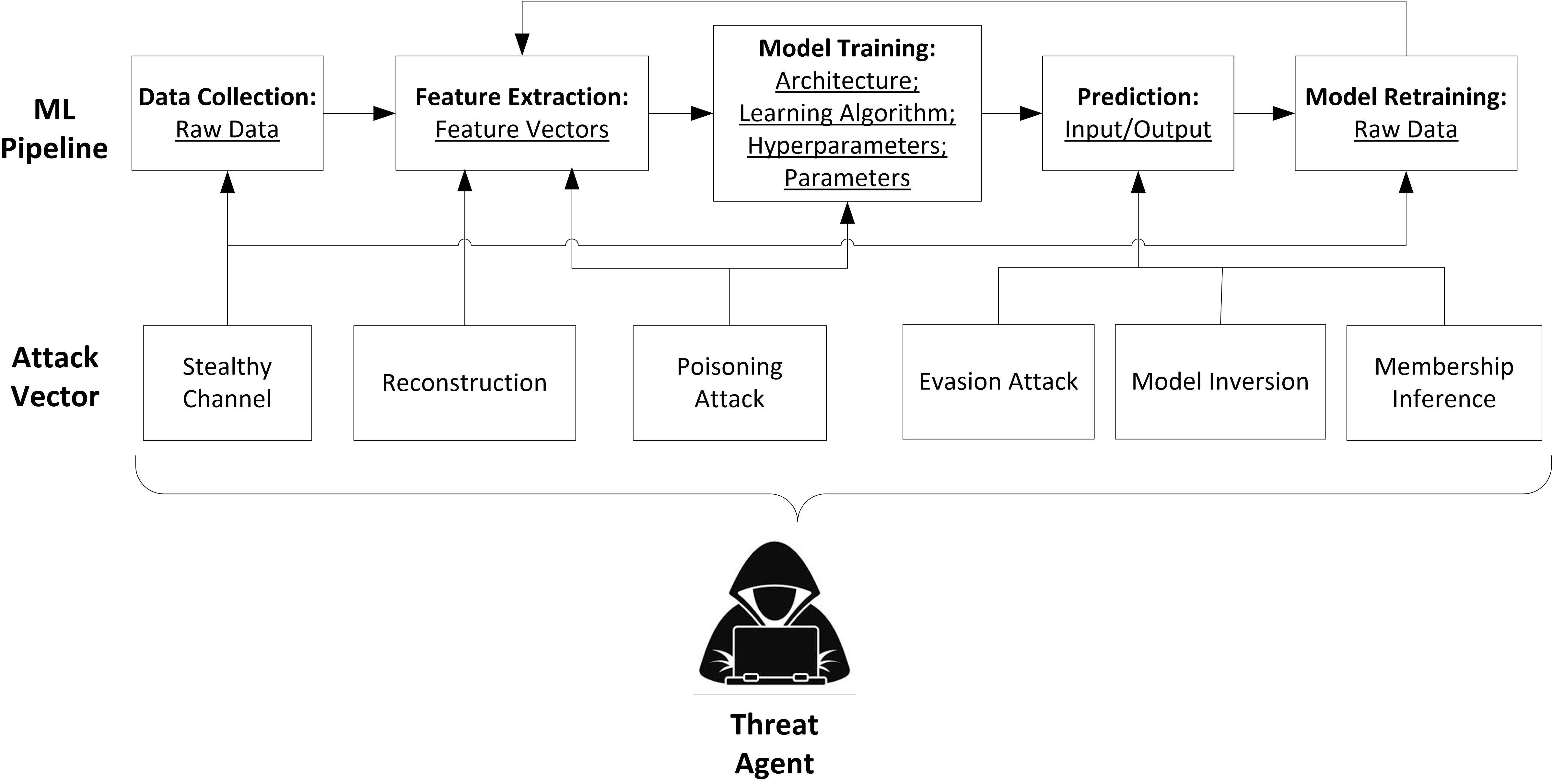}}
\caption{Attack Surface}
\label{fig:AttackSurface}
\end{figure}

\begin{figure}[htbp]
\centerline{\includegraphics[width=0.9\paperwidth]{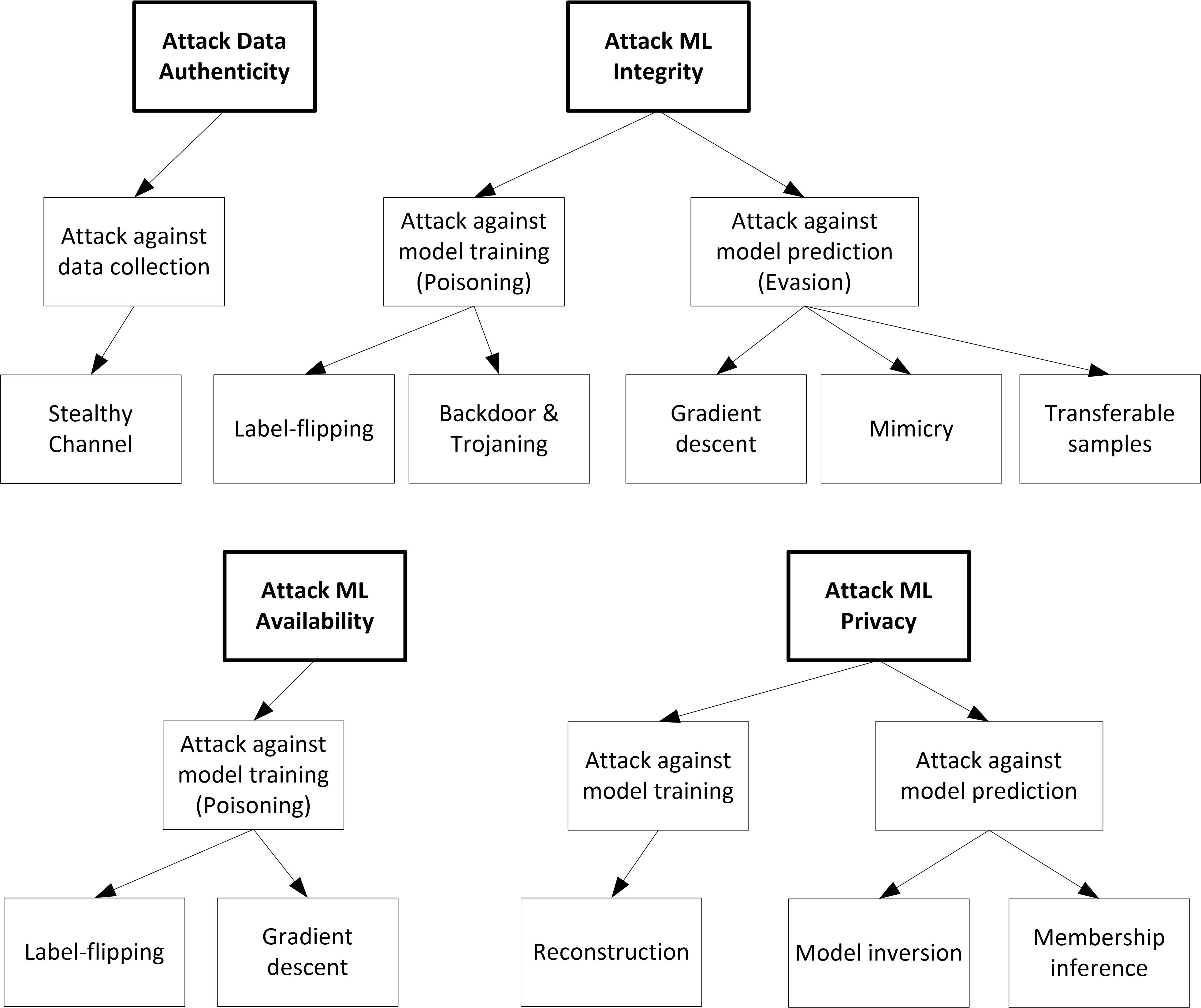}}
\caption{Attack Trees}
\label{fig:AttackTrees}
\end{figure}

\begin{figure}[htbp]
\centerline{\includegraphics[width=0.9\paperwidth]{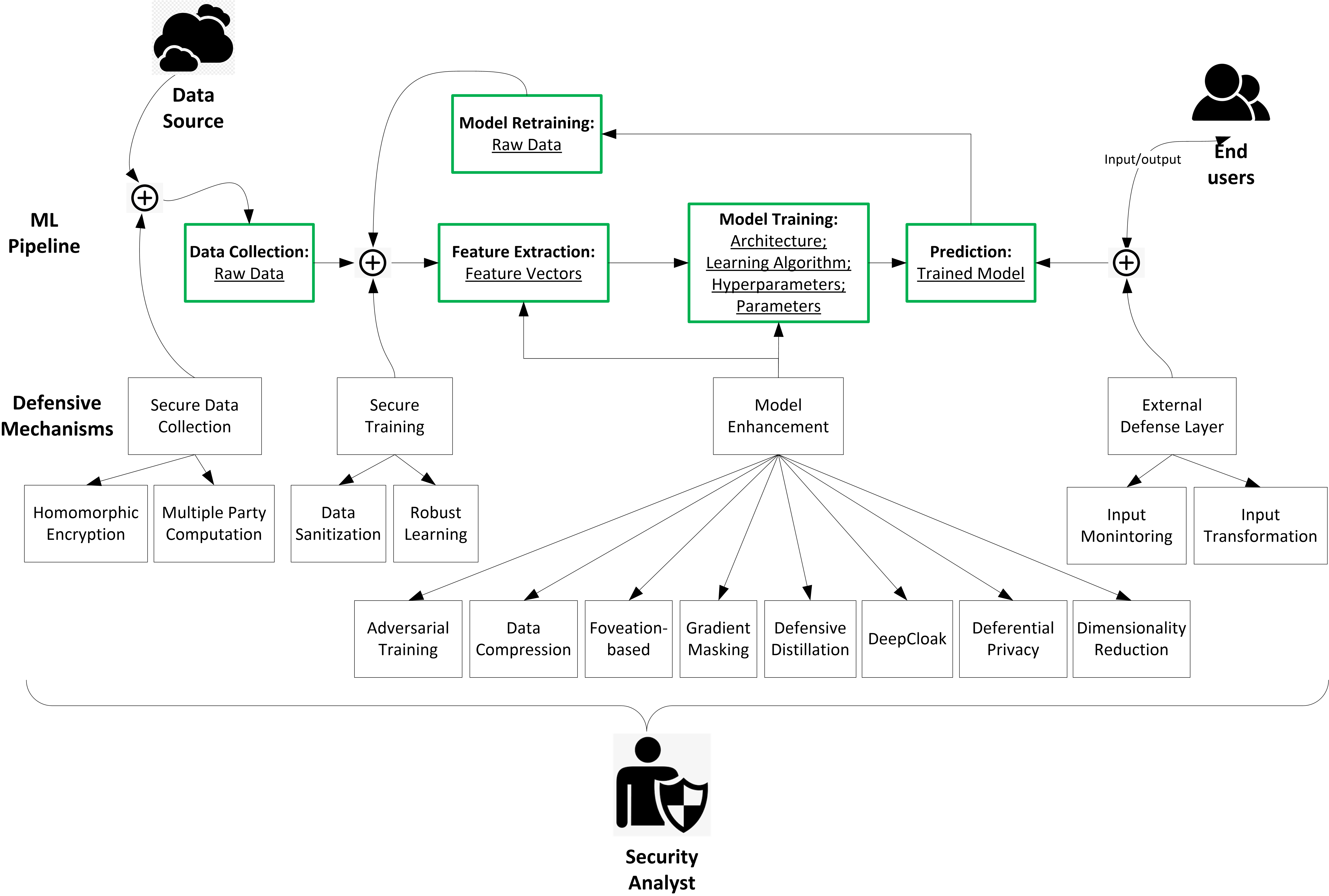}}
\caption{Security Design}
\label{fig:SecurityDesign}
\end{figure}

\end{document}